\documentclass{article}
\usepackage[preprint]{neurips_2026}

\usepackage[utf8]{inputenc} 
\usepackage[T1]{fontenc}    
\usepackage{hyperref}       
\usepackage{url}            
\usepackage{booktabs}       
\usepackage{amssymb}
\usepackage{amsfonts}       
\usepackage{nicefrac}       
\usepackage{microtype}      
\usepackage{xcolor}         
\usepackage{arydshln}
\usepackage{comment}
\usepackage{graphicx}
\usepackage{amsfonts}       
\usepackage{amsmath}
\usepackage{nicefrac}       
\usepackage{microtype}      
\usepackage{xcolor}         
\usepackage{comment}
\usepackage{graphicx}
\usepackage{subcaption}
\usepackage{tikz}
\usepackage{pgfplots}
\usepackage{pgfplotstable}
\usepgfplotslibrary{groupplots}
\pgfplotsset{compat=1.18}
\usepackage{threeparttable}  
\usepackage{pgfplots}
\pgfplotsset{compat=1.18}
\usetikzlibrary{plotmarks,calc}
\usepackage{enumitem}
\usepackage{multirow}
\usepackage{colortbl}   
\usepackage{bm}         
\usepackage{makecell}   
\usepackage{pifont}     

\usepackage{placeins}                      
\providecommand{\shortcite}[1]{\citep{#1}} 
\newcommand{\hl}[1]{#1}                    
\providecommand{\B}[1]{#1}                 

\definecolor{oiBlue}{RGB}{0,114,178}
\definecolor{oiOrange}{RGB}{230,159,0}
\definecolor{oiGreen}{RGB}{0,158,115}
\definecolor{oiRed}{RGB}{213,94,0}
\definecolor{oiPurple}{RGB}{204,121,167}
\definecolor{oiBrown}{RGB}{150,90,50}

\title{WinoTS: Wavelet-based Self-Distillation for Time Series Models}

\author{
Noam Major \quad
Kathy Razmadze \quad
Yoli Shavit \\[0.5em]
Bar-Ilan University, Ramat-Gan, Israel
}

\begin{document}

\maketitle

\begin{abstract}
Self-supervised pre-training of time series models is currently dominated by next-token prediction and reconstruction objectives. In continuous-valued domains, these paradigms often waste model capacity on high-frequency, point-wise noise at the expense of learning invariant structure. While invariance-based self-distillation has proven highly effective in computer vision, its application to temporal data remains largely underexplored. Effectively adapting such methods to time series requires carefully designed augmentations: spatial operations like cropping can shift the timing of repeating cycles or distort the signal, while basic jittering may provide limited variation. We introduce Wavelet-based self-distillation for time series (WinoTS), an invariance-based pre-training paradigm designed specifically for temporal signals. At its core, WinoTS leverages time-frequency augmentations to construct multi-scale structural views without distorting underlying signal dynamics. Across extensive evaluations, WinoTS outperforms state-of-the-art baselines in long-term forecasting, cross-domain zero-shot transfer, and unsupervised anomaly detection. Notably, linear probing on frozen WinoTS representations frequently surpasses fully supervised models trained from scratch. Systematic ablations demonstrate that WinoTS is a flexible, architecture-agnostic framework yielding gains across time series backbones, and establish that time-frequency transformations provide a principled alternative to vision-style spatial augmentations.
\end{abstract}

\section{Introduction}
Self-supervised learning (SSL) has become the primary driver for learning general-purpose representations across modalities~\citep{devlin2019bert,chen2020simple,caron2021dino,bommasani2021opportunities}. In computer vision, joint-embedding self-distillation frameworks such as DINO~\citep{caron2021dino} construct semantic representations by enforcing consistency across augmented views using a momentum teacher. These methods operate on the principle of semantic invariance: a sample's core structure should remain recognizable across label-preserving transformations. However, the efficacy of self-distillation depends on identifying augmentations that induce meaningful invariance without destroying structural integrity.

Currently, self-supervised pre-training for time series remains heavily dominated by generative paradigms, specifically Next Token Prediction (NTP)~\cite{chronos,timesfm} and Masked Auto Encoding (MAE)~\cite{patch_tst,timesiam}. In continuous-valued domains, however, these objectives often force models to dedicate significant capacity to predicting high-frequency, point-wise noise rather than capturing underlying structural invariants. While joint-embedding frameworks based on semantic invariance offer a compelling alternative, their application to temporal data remains largely underexplored.

A central challenge in extending invariance-based self-supervision to time series lies in the design of effective data augmentations. In computer vision, operations such as localized cropping or color jitter alter surface appearance while reliably preserving object identity. Applying analogous transformations to temporal sequences, however, can disrupt fundamental signal properties. For example, cropping a segment from a multi-cycle sequence, such as an electrocardiogram (ECG) heartbeat or an industrial sensor trace, can truncate recurring seasonal patterns, distort global trends, or shift phase alignment. Conversely, basic point-wise operations like Gaussian noise or minor jittering alter individual values without inducing meaningful structural variation, which may yield a weak self-supervisory signal. Effective adaptation therefore requires augmentation strategies that create semantic variation while explicitly preserving the structural and temporal integrity of the underlying signal.

To bridge this gap, we introduce Wavelet-based self-distillation for time series (\textsc{WinoTS}), an invariance-based pre-training paradigm designed specifically for continuous temporal signals. \textsc{WinoTS} leverages the Discrete Wavelet Transform (DWT) to establish a principled augmentation mechanism localized simultaneously in time and frequency. Given an input window, \textsc{WinoTS} decomposes each channel into multi-resolution approximation and detail components to construct asymmetric, full-length views without altering sequence length or discarding temporal context. Specifically, low-frequency approximation coefficients (capturing macro trends and seasonality) remain intact across views, while high-frequency detail subspaces are asymmetrically perturbed: "easy" views smooth these details via soft-thresholding, whereas "hard" views corrupt them with controlled noise. The student network is then optimized to match the teacher's centered and sharpened output distribution across augmented view pairs through a standard DINO projection head. To avoid overfitting to the boundary behaviors or phase characteristics of a single basis, \textsc{WinoTS} stochastically samples wavelet functions across the Daubechies, Symlets, and Coiflets families~\cite{daubechies1988orthonormal,daubechies}.

We evaluate \textsc{WinoTS} through a comprehensive suite of experiments spanning 13 standard forecasting benchmarks, 12 cross-domain transfer scenarios, and 5 multivariate anomaly detection datasets. Across these tasks, \textsc{WinoTS} enhances downstream representation quality, achieving the top average rank on in-domain forecasting and outperforming direct supervised baselines as well as specialized SSL models like TimeSiam~\cite{timesiam} and TS2Vec~\cite{ts2vec}. Under linear probing, frozen \textsc{WinoTS} features demonstrate strong linear separability, frequently surpassing models trained end-to-end from scratch. Furthermore, controlled ablations confirm that wavelet-based view generation provides a principled alternative to aggressive vision primitives (e.g., cropping), delivers competitive performance relative to generative objectives such as MAE and NTP, and yields error reductions across diverse backbone architectures.

In summary, our key contributions are:
\begin{itemize}
    \item We introduce \textsc{WinoTS}, an invariance-based self-distillation paradigm tailored for continuous temporal signals that addresses the limitations of generative pre-training by optimizing for structural invariants rather than point-wise noise prediction.
     \item We propose a multi-resolution wavelet view generator leveraging the Discrete Wavelet Transform (DWT) that preserves low-frequency trend and seasonal components while asymmetrically perturbing high-frequency detail coefficients via soft-thresholding and noise corruption, constructing context-preserving view pairs for self-distillation.
    \item We demonstrate across 13 forecasting benchmarks, 12 cross-domain transfer tasks, and 5 multivariate anomaly detection datasets that \textsc{WinoTS} achieves superior performance over state-of-the-art baselines. Controlled ablations confirm that wavelet-based view generation provides a principled alternative to vision primitives, rivals generative objectives, and delivers gains across diverse time series backbones.
\end{itemize}
\section{Related Work}
\paragraph{Generative vs. Latent-Alignment Pre-training for Time Series Models.}
Self-supervised pre-training for temporal data broadly falls into two primary paradigms: generative modeling and latent alignment. Generative approaches learn representations by predicting, denoising, or reconstructing raw temporal signals in the target data domain. These include masked autoencoders~\cite{patch_tst}, autoregressive next-token prediction~\cite{chronos,timesfm}, diffusion frameworks~\cite{timedart}, and past-to-present target-domain reconstruction via Siamese networks~\cite{timesiam}. Although effective for forecasting, these point-level objectives force models to dedicate significant capacity to predicting fine-grained numerical values and high-frequency noise at the expense of capturing broader invariant structure~\cite{major2026quantifying}. Conversely, latent-alignment paradigms directly optimize representation space by enforcing consistency across different views of the same signal without signal-space decoders. Early work relied primarily on contrastive learning and Siamese instance matching~\cite{cpc,ts2vec}, whereas recent advances have embraced non-contrastive self-distillation and momentum teacher--student objectives inspired by DINO~\cite{pieper2023selfdistilled,himtm,moakher2026utica}. However, existing latent-alignment approaches predominantly rely on basic spatial masking or cropping, and their empirical validation remains largely constrained to narrow benchmark sets or single architectural backbones.

\paragraph{Time-Series Augmentations for Latent Alignment.}
The representation quality of latent-alignment frameworks depends fundamentally on the choice of view generation. Existing methods predominantly rely on spatial or time-domain transformations, such as cropping, temporal jittering, or channel masking~\cite{ts2vec,moakher2026utica}. However, these vision-inspired operations can severely disrupt temporal dynamics: aggressive cropping removes essential seasonal and macro-trend context, while point-wise noise degrades high-frequency phase and amplitude relationships. Recent studies have highlighted this trade-off, analyzing the precision--invariance spectrum under standardized benchmarks~\cite{major2026quantifying}. Rather than applying spatial or heuristic transformations in the time domain, \textsc{WinoTS} introduces a principled wavelet-based view generation framework that operates natively across time and frequency, constructing multi-scale views that induce invariance while strictly preserving signal dynamics.
\section{Method}
\label{sec:method}
We present \textsc{WinoTS}, a self-supervised pre-training framework that adapts joint-embedding self-distillation to temporal data via multi-resolution time-frequency view generation. Instead of manipulating time-domain samples directly, \textsc{WinoTS} maps input sequences into wavelet coefficient space via the Discrete Wavelet Transform (DWT), constructing asymmetric, full-length views that preserve macroscopic trend and seasonal context while selectively perturbing high-frequency details. In the following subsections, we formalize the wavelet-based view generation mechanism, detail the asymmetric teacher--student self-distillation objective, and describe our stochastic basis sampling scheme.

\subsection{Background}
\label{sub:background}

We first provide the necessary mathematical background on self-distillation and multi-resolution signal representations required to ground our framework.

\paragraph{Self-Distillation with No Labels (DINO).}
Originally introduced in computer vision, DINO \cite{caron2021dino} serves as a general joint-embedding framework for learning semantic representations without negative pairs or labels. The framework consists of two networks sharing identical structural topologies: a student network $g_{\theta_s, \phi_s}$ optimized via gradient descent, and a teacher network $g_{\theta_t, \phi_t}$ whose parameters are updated as an exponential moving average (EMA) of the student weights. Each network is parameterized as the composition of an encoder backbone $f_{\theta}$ and a non-linear projection head $h_{\phi}$:
\begin{equation}
g_{\theta,\phi} = h_{\phi} \circ f_{\theta}.
\end{equation}
Given two different augmented views of an input sample, denoted $x_s$ and $x_t$, the student and teacher process them to produce output logits $z_s = g_{\theta_s, \phi_s}(x_s)$ and $z_t = g_{\theta_t, \phi_t}(x_t)$. These logits are converted into probability distributions using a softmax operator. The student's predictive distribution is parameterized by a temperature hyperparameter $\tau_s$:
\begin{equation}
P_s^{(i)}(z_s) = \frac{\exp(z_s^{(i)}/\tau_s)}{\sum_j \exp(z_s^{(j)}/\tau_s)}.
\end{equation}
To prevent representation collapse across channels, the teacher's target distribution is centered by a running mean vector $c$ and sharpened by a distinct target temperature $\tau_t < \tau_s$:
\begin{equation}
P_t^{(i)}(z_t) = \frac{\exp((z_t^{(i)}-c^{(i)})/\tau_t)}{\sum_j \exp((z_t^{(j)}-c^{(j)})/\tau_t)}.
\end{equation}
The framework is optimized by minimizing the cross-entropy loss between the teacher's centered and sharpened target distribution and the student's distribution:
\begin{equation}
\mathcal{L}_{\mathrm{DINO}} = - \sum_i P_t^{(i)}(z_t) \log P_s^{(i)}(z_s).
\end{equation}
Gradients are backpropagated strictly through the student network. After each optimization step, the teacher's parameters are updated via a momentum schedule: $\theta_t \leftarrow \lambda \theta_t + (1-\lambda)\theta_s$ and $\phi_t \leftarrow \lambda \phi_t + (1-\lambda)\phi_s$. 

\paragraph{Discrete Wavelet Decomposition.}
The Discrete Wavelet Transform (DWT) provides a standard mathematical framework for analyzing non-stationary signals across multiple resolutions simultaneously. Formally, let $x \in \mathbb{R}^{T}$ denote a single-channel temporal sequence. A $J$-level DWT parameterized by a chosen wavelet basis $w$ recursively decomposes the signal into a low-frequency approximation band and multiple high-frequency detail bands. This decomposition is achieved by applying a low-pass filter $h$ and a high-pass filter $g$, followed by dyadic downsampling at each scale:
\begin{equation}
\begin{split}
a_j &= (h \star a_{j-1})\downarrow 2, \\
d_j &= (g \star a_{j-1})\downarrow 2, \qquad j=1,\dots,J,
\end{split}
\end{equation}
where $a_0 = x$. This forward projection maps the raw signal into an orthogonal wavelet space $\mathcal{W}_w x = (a_J, d_J, \dots, d_1)$. For an orthonormal basis, the original signal can be perfectly recovered via the inverse reconstruction formula:
\begin{equation}
x = \sum_k a_{J,k}\phi_{J,k} + \sum_{j=1}^{J}\sum_k d_{j,k}\psi_{j,k},
\end{equation}
where $\phi_{J,k}$ and $\psi_{j,k}$ represent the scaling and wavelet functions, respectively. The first term yields the coarse approximation projection $P_{V_J}x$, while the second term captures multi-scale localized temporal fluctuations in the orthogonal complement space $V_J^\perp$.

\subsection{\textsc{WinoTS}: Self-Distillation with Wavelet-based View Construction}
\label{sub:wino_construction}
\begin{figure}[t]
    \centering
    \makebox[\textwidth][c]{%
        \includegraphics[width=1.15\textwidth]{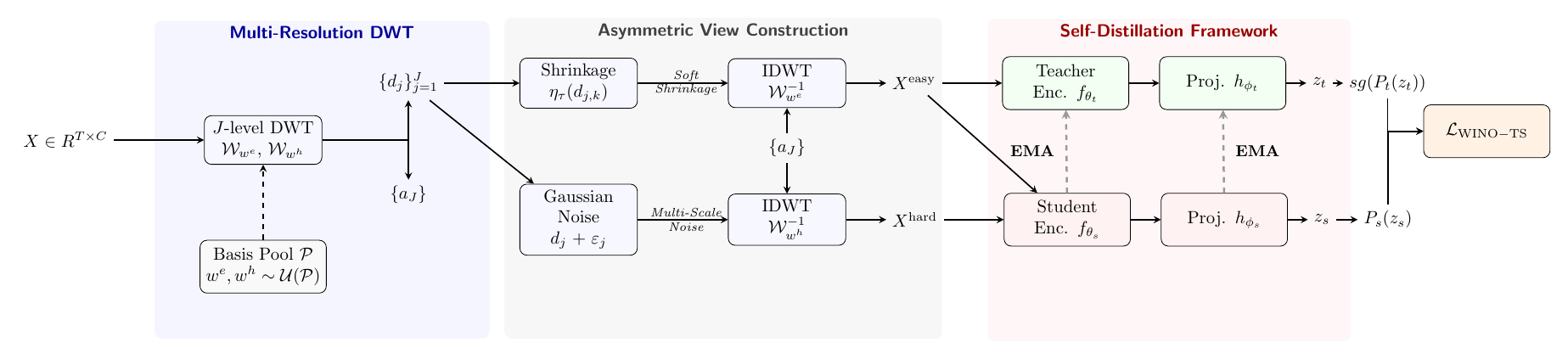}%
    }
    \caption{
Architectural pipeline of \textsc{WinoTS}. The wavelet generator produces $Q$ easy and $V$ hard views. The teacher processes only easy views, while the student processes both. Each teacher easy view is matched to all student views except its identical easy counterpart; every hard view is matched to all teacher views.
}
    \label{fig:winots}
\end{figure}

\textsc{WinoTS} adapts the general DINO framework to time series by introducing a novel view-generation strategy natively tailored to the temporal domain. Instead of relying on spatial or spatial-spectral approximations, our framework exploits the time-frequency locality of the DWT to construct asymmetric, full-length temporal views (Fig.~\ref{fig:winots}).

\paragraph{Easy View Generation.}
\textsc{WinoTS} constructs $Q$ easy views $\mathcal{E}(X)=\{X_q^{\mathrm{easy}}\}_{q=1}^{Q}$ by applying nonlinear soft-thresholding to the high-frequency detail bands while leaving each view's low-frequency approximation coefficients unchanged.

For a given easy view, the modified coefficients are reconstructed in the time domain using the inverse discrete wavelet transform (IDWT), $\mathcal{W}_w^{-1}$:
\begin{equation}
x^{\mathrm{easy}} = \mathcal{W}_w^{-1} \big( a_J, \eta_{\tau_J}(d_J), \dots, \eta_{\tau_1}(d_1) \big).
\end{equation}
Here, the approximation component $a_J$ remains unchanged, while the detail shrinkage operator $\eta_{\tau}$ is defined independently for each band using its maximum absolute coefficient:
\begin{equation}
\eta_{\tau}(d)= \operatorname{sign}(d)\max(|d|-\tau,0), \qquad \tau_j=\rho\max_k |d_{j,k}|.
\end{equation}
This operation sets detail coefficients below $\tau_j$ to zero and attenuates larger coefficients, retaining dominant localized fluctuations without discarding the entire high-pass subspace. The relationship between this operator and classical wavelet shrinkage is discussed in the Supplementary Material, and sensitivity to the shrinkage parameter is evaluated in Table~\ref{tab:rho-ablation-appendix}.

\paragraph{Hard View Generation.}
\textsc{WinoTS} further generates $V$ hard views $\{X_v^{\mathrm{hard}}\}_{v=1}^{V}$. A hard view is constructed by injecting controlled Gaussian noise into the multi-scale detail bands while keeping the approximation coefficients $a_J$ unchanged, followed by IDWT reconstruction:
\begin{equation}
x^{\mathrm{hard}} = \mathcal{W}_w^{-1} \big( a_J, d_J+\varepsilon_J, \dots, d_1+\varepsilon_1 \big), \qquad \varepsilon_{j,k} \sim \mathcal{N}(0,s^2).
\end{equation}
For a multivariate hard view, the perturbation scale is sampled independently for each channel. Specifically, for hard view $v$ and channel $c$,
\begin{equation}
s_{v,c} \sim \mathcal{U}(s_{\min},s_{\max}), \qquad \varepsilon_{v,j,k,c} \sim \mathcal{N}(0,s_{v,c}^{2}).
\end{equation}
The sampled scale $s_{v,c}$ is shared across all detail levels of channel $c$ within that hard view. The effect of the perturbation design is further analyzed through the augmentation and transform ablations in Tables~\ref{tab:app-augmentation-ablation} and \ref{tab:app-wavelet-transform}.

\paragraph{Multi-Resolution Basis Randomization.}
A single wavelet basis introduces specific mathematical artifacts rooted in its unique phase alignment, compact support limits, and boundary behaviors. To prevent the model from overfitting to these basis-specific traits, \textsc{WinoTS} samples distinct wavelet bases independently for the teacher and student views from an orthogonal pool $\mathcal{P}$:
\begin{equation}
w^{\mathrm{easy}}, w^{\mathrm{hard}} \overset{\mathrm{i.i.d.}}{\sim} \mathcal{U}(\mathcal{P}),
\end{equation}
where the pool consists of six compactly supported wavelets spanning three distinct mathematical families: $\mathcal{P} = \{ \mathrm{sym}4, \mathrm{sym}6, \mathrm{sym}8, \mathrm{db}4, \mathrm{db}6, \mathrm{coif}2 \}.$
These families introduce diverse signal processing properties into the augmentation pipeline. Daubechies (\texttt{db}) wavelets provide a minimum-phase asymmetric design, Symlets (\texttt{sym}) optimize for near-symmetric phase behavior to minimize phase distortion, and Coiflets (\texttt{coif}) balance symmetry by enforcing vanishing moment constraints on both the scaling and wavelet functions. The properties of the sampled wavelet families are summarized in Supplementary Table A1. The impact of alternative wavelet families, decomposition depths, and transform variants is evaluated in Supplementary Tables~\ref{tab:app-wavelet-transform}--\ref{tab:wavelet-pool-ablation-appendix}.

\paragraph{Architecture-Agnostic Integration.}
\label{sub:multivariate}
For a multivariate input window $X \in \mathbb{R}^{T \times C}$, \textsc{WinoTS} applies the wavelet transform channel-wise. Within a single view, all channels share the same sampled wavelet basis $w$. For each hard view $v$, however, every channel $c$ independently samples a perturbation scale $s_{v,c} \sim \mathcal{U}(s_{\min}, s_{\max})$ shared across its detail levels, along with an independent noise realization $\varepsilon_{v,j,k,c} \sim \mathcal{N}(0, s_{v,c}^{2})$. This design preserves cross-channel view coordination while preventing artificial high-frequency co-dependence.

Crucially, because view generation operates in the wavelet coefficient domain prior to full-length reconstruction, the sequence length $T$ remains strictly invariant across all views. This makes \textsc{WinoTS} natively model-agnostic: any time series backbone $f_\theta$ designed for uniform sequence lengths can be seamlessly embedded without modifying its structural parameters or positional encodings (empirically demonstrated across backbone families in Supplementary Table~\ref{tab:backbone-ablation}).

\paragraph{The \textsc{WinoTS} Objective.}
For an input window $X$, let $\mathcal{E}(X) = \{X_q^{\mathrm{easy}}\}_{q=1}^{Q}$ and $\mathcal{H}(X) = \{X_v^{\mathrm{hard}}\}_{v=1}^{V}$ denote the sets of $Q$ easy views and $V$ hard views, respectively. The momentum teacher processes exclusively easy views from $\mathcal{E}(X)$, whereas the student network processes views from both $\mathcal{E}(X)$ and $\mathcal{H}(X)$.

Let $P_{t,q}^{\mathrm{easy}}$ denote the teacher's target distribution for easy view $q$, and let $P_{s,u}^{\mathrm{easy}}$ and $P_{s,v}^{\mathrm{hard}}$ denote the student's output distributions for easy view $u$ and hard view $v$. Following standard cross-view self-distillation, valid easy-teacher/easy-student pairs exclude identical view realizations:
\begin{equation}
\mathcal{I}_{\mathrm{ee}} = \left\{ (q,u): 1\leq q,u\leq Q,\; q\neq u \right\},
\end{equation}
whereas all easy-teacher/hard-student pairs are valid:
\begin{equation}
\mathcal{I}_{\mathrm{eh}} = \left\{ (q,v): 1\leq q\leq Q,\; 1\leq v\leq V \right\}.
\end{equation}

The \textsc{WinoTS} objective minimizes cross-entropy across all valid view combinations:
\begin{equation}
\label{eq:wino-loss-final}
\begin{split}
\mathcal{L}_{\text{\textsc{WinoTS}}} = \frac{1}{|\mathcal{I}_{\mathrm{ee}}| + |\mathcal{I}_{\mathrm{eh}}|} \Bigg[ & \sum_{(q,u)\in\mathcal{I}_{\mathrm{ee}}} H\!\left( \operatorname{sg}\left[P_{t,q}^{\mathrm{easy}}\right], P_{s,u}^{\mathrm{easy}}\right) \\
+ & \sum_{(q,v)\in\mathcal{I}_{\mathrm{eh}}} H\!\left( \operatorname{sg}\left[P_{t,q}^{\mathrm{easy}}\right], P_{s,v}^{\mathrm{hard}}\right) \Bigg],
\end{split}
\end{equation}
where $\operatorname{sg}[\cdot]$ denotes the stop-gradient operator. Here, $|\mathcal{I}_{\mathrm{ee}}|=Q(Q-1)$ and $|\mathcal{I}_{\mathrm{eh}}|=QV$, yielding an average over $Q(Q+V-1)$ valid pair combinations (alternative pairing configurations are evaluated in Supplementary Table~\ref{tab:view-pairing-ablation-appendix}).

Crucially, because wavelet bases $w^{\mathrm{easy}}$ and $w^{\mathrm{hard}}$ are sampled independently per view pair, the teacher and student evaluate input representations across distinct time-frequency coordinate frames. This basis mismatch introduces subtle phase and boundary shifts across shared low-frequency bands, forcing the encoder to learn representations invariant to coordinate choices and cross-basis phase leakage. We detail the projection head specifications, teacher momentum schedules, and complete hyperparameter setups in Supplementary Tables~\ref{tab:app-wavelet-pool} and~\ref{tab:app-pretrain-hparams}.

\begin{table*}[h!]
\centering
\caption{Average in-domain forecasting over horizons
$\{96,192,336,720\}$ with context length 336 (lower is better).
Red and blue indicate best and second-best results.
\textsc{WinoTS} uses full fine-tuning; \textsc{WinoTS}-LP freezes the backbone and
trains only the forecasting head. TimeSiam and TS2Vec are
self-supervised; the remaining baselines are supervised.}

\label{tab:forecasting-main}
\scriptsize
\setlength{\tabcolsep}{2.2pt}
\resizebox{\textwidth}{!}{%
\begin{tabular}{lcccccccccccccccccccccccccc}
\toprule
 & \multicolumn{4}{c}{\textbf{Ours}} & \multicolumn{4}{c}{\textbf{Self-Supervised}} & \multicolumn{18}{c}{\textbf{Supervised}} \\
\cmidrule(lr){2-5}\cmidrule(lr){6-9}\cmidrule(lr){10-27}
\textbf{Dataset} & \multicolumn{2}{c}{\textsc{WinoTS}} & \multicolumn{2}{c}{\textsc{WinoTS}-LP} & \multicolumn{2}{c}{TimeSiam} & \multicolumn{2}{c}{TS2Vec} & \multicolumn{2}{c}{TimeMixer} & \multicolumn{2}{c}{TimeBase} & \multicolumn{2}{c}{SparseTSF} & \multicolumn{2}{c}{PatchTST} & \multicolumn{2}{c}{DLinear} & \multicolumn{2}{c}{iTransformer} & \multicolumn{2}{c}{FEDformer} & \multicolumn{2}{c}{TimesNet} & \multicolumn{2}{c}{Autoformer} \\
 & \multicolumn{2}{c}{(ours)} & \multicolumn{2}{c}{(ours)} & \multicolumn{2}{c}{\begin{tabular}[t]{@{}c@{}}\citeauthor{timesiam}\\ {[}\citeyear{timesiam}{]}\end{tabular}} & \multicolumn{2}{c}{\begin{tabular}[t]{@{}c@{}}\citeauthor{ts2vec}\\ {[}\citeyear{ts2vec}{]}\end{tabular}} & \multicolumn{2}{c}{\begin{tabular}[t]{@{}c@{}}\citeauthor{wang2024timemixer}\\ {[}\citeyear{wang2024timemixer}{]}\end{tabular}} & \multicolumn{2}{c}{\begin{tabular}[t]{@{}c@{}}\citeauthor{timebase}\\ {[}\citeyear{timebase}{]}\end{tabular}} & \multicolumn{2}{c}{\begin{tabular}[t]{@{}c@{}}\citeauthor{sparsetsf}\\ {[}\citeyear{sparsetsf}{]}\end{tabular}} & \multicolumn{2}{c}{\begin{tabular}[t]{@{}c@{}}\citeauthor{patch_tst}\\ {[}\citeyear{patch_tst}{]}\end{tabular}} & \multicolumn{2}{c}{\begin{tabular}[t]{@{}c@{}}\citeauthor{dlinear}\\ {[}\citeyear{dlinear}{]}\end{tabular}} & \multicolumn{2}{c}{\begin{tabular}[t]{@{}c@{}}\citeauthor{liu2023itransformer}\\ {[}\citeyear{liu2023itransformer}{]}\end{tabular}} & \multicolumn{2}{c}{\begin{tabular}[t]{@{}c@{}}\citeauthor{fedformer}\\ {[}\citeyear{fedformer}{]}\end{tabular}} & \multicolumn{2}{c}{\begin{tabular}[t]{@{}c@{}}\citeauthor{wu2023timesnet}\\ {[}\citeyear{wu2023timesnet}{]}\end{tabular}} & \multicolumn{2}{c}{\begin{tabular}[t]{@{}c@{}}\citeauthor{autoformer}\\ {[}\citeyear{autoformer}{]}\end{tabular}} \\
\cmidrule(lr){2-3}\cmidrule(lr){4-5}\cmidrule(lr){6-7}\cmidrule(lr){8-9}\cmidrule(lr){10-11}\cmidrule(lr){12-13}\cmidrule(lr){14-15}\cmidrule(lr){16-17}\cmidrule(lr){18-19}\cmidrule(lr){20-21}\cmidrule(lr){22-23}\cmidrule(lr){24-25}\cmidrule(lr){26-27}
 & MSE & MAE & MSE & MAE & MSE & MAE & MSE & MAE & MSE & MAE & MSE & MAE & MSE & MAE & MSE & MAE & MSE & MAE & MSE & MAE & MSE & MAE & MSE & MAE & MSE & MAE \\
\midrule
ETTh1 & \textcolor{blue}{0.411} & \textcolor{blue}{0.423} & 0.416 & \textcolor{blue}{0.423} & 0.426 & 0.443 & 0.817 & 0.669 & 0.436 & 0.444 & \textcolor{red}{0.409} & \textcolor{red}{0.415} & 0.423 & 0.430 & 0.437 & 0.438 & 0.465 & 0.469 & 0.448 & 0.446 & 0.484 & 0.490 & 0.471 & 0.465 & 0.568 & 0.518 \\
ETTh2 & \textcolor{red}{0.347} & \textcolor{blue}{0.386} & \textcolor{red}{0.347} & \textcolor{red}{0.385} & 0.362 & 0.402 & 1.957 & 1.108 & \textcolor{blue}{0.349} & 0.391 & 0.358 & 0.399 & 0.372 & 0.406 & 0.380 & 0.407 & 0.458 & 0.462 & 0.379 & 0.397 & 0.419 & 0.460 & 0.408 & 0.405 & 0.526 & 0.505 \\
ETTm1 & \textcolor{red}{0.346} & \textcolor{red}{0.379} & 0.362 & 0.386 & \textcolor{blue}{0.348} & \textcolor{blue}{0.384} & 0.670 & 0.583 & 0.362 & 0.388 & 0.367 & 0.386 & 0.368 & 0.394 & 0.393 & 0.401 & 0.377 & 0.396 & 0.408 & 0.410 & 0.441 & 0.459 & 0.403 & 0.412 & 0.665 & 0.545 \\
ETTm2 & \textcolor{red}{0.250} & \textcolor{red}{0.308} & \textcolor{blue}{0.252} & \textcolor{red}{0.308} & 0.260 & 0.321 & 0.912 & 0.676 & 0.261 & 0.321 & 0.260 & \textcolor{blue}{0.317} & 0.264 & 0.320 & 0.282 & 0.327 & 0.317 & 0.371 & 0.291 & 0.334 & 0.329 & 0.377 & 0.296 & 0.332 & 0.356 & 0.401 \\
Weather & \textcolor{red}{0.224} & \textcolor{red}{0.262} & 0.238 & 0.272 & 0.228 & \textcolor{blue}{0.264} & 1.021 & 0.727 & \textcolor{blue}{0.226} & 0.265 & 0.254 & 0.290 & 0.235 & 0.277 & 0.266 & 0.284 & 0.249 & 0.300 & 0.261 & 0.281 & 0.311 & 0.363 & 0.259 & 0.287 & 0.331 & 0.373 \\
Electricity & \textcolor{blue}{0.163} & \textcolor{blue}{0.254} & 0.167 & 0.260 & \textcolor{red}{0.158} & \textcolor{red}{0.249} & 0.373 & 0.451 & \textcolor{blue}{0.163} & 0.255 & 0.176 & 0.261 & 0.166 & 0.257 & 0.365 & 0.299 & 0.173 & 0.275 & 0.183 & 0.274 & 0.256 & 0.360 & 0.201 & 0.298 & 0.321 & 0.392 \\
Exchange & 	0.377 & 0.407 & 0.397 & 0.419 & 0.412 & 0.431 & 0.831 & 0.639 & 0.458 & 0.454 & 0.450 & 0.461 & 0.440 & 0.458 & 0.365 & \textcolor{blue}{0.405} & \textcolor{red}{0.317} & 0.414 & \textcolor{blue}{0.358} & \textcolor{red}{0.404} & 0.825 & 0.675 & 0.424 & 0.449 & 0.793 & 0.656 \\
Solar & \textcolor{blue}{0.204} & \textcolor{blue}{0.258} & 0.260 & 0.315 & 0.213 &  0.273 & 0.274 & 0.365 & 0.213 & 0.277 & 0.292 & 0.307 & \textcolor{red}{0.196} & \textcolor{red}{0.246} & 0.270 & 0.304 & 0.258 & 0.319 & 0.272 & 0.300 & 0.310 & 0.402 & 0.274 & 0.294 & 0.740 & 0.631 \\
Traffic & \textcolor{blue}{0.411} & \textcolor{blue}{0.276} & 0.420 & 0.282 & \textcolor{red}{0.407} & 0.280 & 0.938 & 0.545 & 0.430 & 0.310 & 0.447 & 0.293 & 0.415 & \textcolor{red}{0.267} & 0.471 & 0.304 & 0.451 & 0.320 & 0.479 & 0.326 & 0.640 & 0.391 & 0.638 & 0.335 & 0.745 & 0.445 \\
AQShunyi & 	\textcolor{blue}{0.678} & \textcolor{blue}{0.503} & 0.684 & \textcolor{red}{0.502} & 0.680 & \textcolor{blue}{0.504} & \textcolor{red}{0.672} & 0.515 & 0.686 & 0.504 & 0.705 & 0.518 & 0.695 & 0.518 & 0.775 & 0.529 & 0.695 & 0.527 & 0.780 & 0.528 & 0.799 & 0.562 & 0.754 & 0.525 & 0.799 & 0.563 \\
AQWan & \textcolor{blue}{0.776} & \textcolor{red}{0.494} & 0.784 & \textcolor{blue}{0.496} & 0.778 & \textcolor{blue}{0.495} & \textcolor{red}{0.765} & 0.509 & 0.784 & 0.497 & 0.811 & 0.511 & 0.795 & 0.509 & 0.873 & 0.517 & 0.798 & 0.520 & 0.883 & 0.519 & 0.814 & 0.526 & 0.846 & 0.511 & 0.870 & 0.552 \\
CzeLan & \textcolor{red}{0.223} & \textcolor{red}{0.262} & \textcolor{blue}{0.230} & \textcolor{blue}{0.267} & 0.232 & 0.287 & 0.333 & 0.413 & \textcolor{blue}{0.230} & 0.282 & 0.274 & 0.325 & 0.238 & 0.286 & 0.266 & 0.299 & 0.354 & 0.385 & 0.267 & 0.295 & 0.333 & 0.386 & 0.307 & 0.325 & 0.655 & 0.580 \\
PM2.5 & \textcolor{red}{0.418} & \textcolor{red}{0.420} & 0.421 & \textcolor{blue}{0.421} & \textcolor{blue}{0.419} & 0.426 & 0.428 & 0.478 & 0.424 & 0.429 & 0.445 & 0.450 & 0.423 & 0.450 & 0.426 & 0.426 & 0.423 & 0.467 & 0.428 & 0.427 & 0.455 & 0.493 & 0.438 & 0.434 & 0.475 & 0.471 \\
\bottomrule
\end{tabular}%
}
\end{table*}

\section{Experimental Setup}
\label{sub:setup}
\paragraph{Datasets and evaluation protocols.}We evaluate forecasting on 13 benchmarks: ETTh1, ETTh2, ETTm1,
ETTm2\footnote{\url{https://github.com/zhouhaoyi/ETDataset}},
Weather\footnote{\url{https://www.bgc-jena.mpg.de/wetter}},
Electricity\footnote{\url{https://archive.ics.uci.edu/ml/datasets}},
Traffic\footnote{\url{https://pems.dot.ca.gov/}}, Exchange and Solar Energy
\citep{lai2018modeling}, AQShunyi and AQWan \citep{zhang2017cautionary},
CzeLan \citep{poyatos2020global}, and PM2.5.
We report MSE and MAE averaged over horizons $\{96,192,336,720\}$. For anomaly detection, we follow the TSLib evaluation protocol and
report point-adjusted precision, recall, and F1; complete detector and
evaluation details are provided in
the supplementary material.

\textsc{WinoTS} is evaluated under two downstream adaptation protocols: by default, \textsc{WinoTS} employs full fine-tuning, where all model parameters are updated end-to-end on the downstream task. To evaluate representation quality, we also test under \textit{linear probing} (denoted by \textsc{WinoTS}-LP), where the pre-trained backbone remains frozen and only a task-specific head is optimized. We compare our framework against eleven state-of-the-art baselines for forecasting and seven baselines for anomaly detection. Unless specified otherwise, all methods use their optimal default backbone architecture (when applicable), optimizer configuration, and total number of optimization steps. For \textsc{WinoTS}, this budget is divided between self-supervised pre-training and downstream adaptation, whereas supervised baselines use their default optimal optimization budget entirely for supervised training.

Unless otherwise stated, \textsc{WinoTS} utilizes in-domain self-supervised pre-training, optimizing on the unlabeled training split of the downstream dataset prior to task-specific adaptation. For cross-domain zero-shot transfer, pre-training and task configuration are performed strictly on the source dataset, and the model is evaluated directly on the target dataset without any target-domain parameter updates.

\paragraph{Implementation Details.} \textsc{WinoTS} pre-trains a TimeMixer backbone using symmetric DWT
boundary extension, decomposition depth $J{=}3$, shrinkage ratio
$\rho{=}0.6$, and the wavelet pool listed in Supplementary Table~A1.
For its longest filter ($\mathrm{sym}8$, $L{=}16$), this depth
satisfies $T\geq2^J(L-1)=120$. Each input yields $Q{=}1$ easy view
and $V{=}1$ hard view. Both views are forwarded through the student,
but the same-index easy--easy pair is excluded; hence
$\mathcal{I}_{\mathrm{ee}}=\varnothing$, and the reported loss reduces
to
$H(\operatorname{sg}[P_t^{\mathrm{easy}}],
P_s^{\mathrm{hard}})$.
Alternative pairing rules and view counts are evaluated in
Supplementary Tables~\ref{tab:view-pairing-ablation-appendix} and \ref{tab:app-hardcrop-sweep}. Inputs are standardized
channel-wise using training-split statistics. For each hard view, the
noise scale is sampled independently per channel from
$\mathcal{U}(0.2,0.5)$ and shared across that channel's detail levels.

The DINO head is a three-layer MLP with hidden width $2048$, a
$256$-dimensional $\ell_2$-normalized bottleneck, and a
weight-normalized output layer with $K{=}1024$; the choice of $K$ is
evaluated in Supplementary Table~A4. We use $\tau_s{=}0.1$,
$\tau_t{=}0.04$, center momentum $0.9$, and a cosine teacher-EMA
schedule from $0.9996$ to $1.0$. After pre-training, the projection
head is discarded and the EMA teacher backbone is used for downstream
adaptation. Complete pre-training optimization and architecture settings are
reported in Supplementary Table~\ref{tab:app-pretrain-hparams}; the wavelet pool and projection
output-dimension study are provided in Supplementary Tables~\ref{tab:app-wavelet-pool} and \ref{tab:outdim-ablation}.

\paragraph{Architectural and Design Ablations.} 
To isolate the factors driving performance, we evaluate core design choices in Section~\ref{sec:results}, including view generation primitives, loss objectives, and encoder backbones. Comprehensive supplementary evaluations, covering synthetic pre-training, backbone transferability, wavelet family choices, view pairing rules, and seed robustness, are detailed in Supplementary Tables~\ref{tab:app-synthetic} and~\ref{tab:backbone-ablation}--\ref{tab:seed-robustness}.

\section{Results and Discussion}
\label{sec:results}
We systematically evaluate \textsc{WinoTS} across three primary downstream tasks: in-domain forecasting, cross-domain zero-shot transfer, and unsupervised anomaly detection. Through extensive empirical comparisons, we demonstrate that wavelet-based self-distillation consistently outperforms supervised baselines, generalizes across diverse backbone architectures, and offers clear advantages over both spatial edits and generative pre-training objectives.

\subsection{In-Domain Forecasting, Linear Separability, and Backbone Generalization}
\label{sub:res_indomain}
Table~\ref{tab:forecasting-main} reports in-domain forecasting performance across 13 standard benchmarks (complete per-horizon results are provided in Supplementary Table~\ref{tab:forecasting-perhorizon}). We evaluate two downstream protocols using our pre-trained TimeMixer backbone: \textsc{WinoTS} (full end-to-end fine-tuning) and \textsc{WinoTS}-LP (linear probing over the frozen pre-trained backbone). All baseline methods, including the supervised TimeMixer baseline, are fully trained end-to-end from scratch. \textsc{WinoTS} consistently improves upon its supervised TimeMixer counterpart, achieving the top average rank across all evaluated methods. Compared directly against supervised TimeMixer, \textsc{WinoTS} yields clear error reductions on 12 datasets by MSE and all 13 by MAE, indicating that wavelet-based self-distillation captures structural regularities that complement the backbone's native inductive bias. While specialized architectures retain localized advantages on specific benchmarks—such as DLinear on Exchange or SparseTSF on Solar—\textsc{WinoTS} delivers the most consistent gains across the benchmark suite, ranking among the top two performing methods on 12 out of 13 datasets.

To isolate representation quality from fine-tuning dynamics, we evaluate the frozen backbone under \textsc{WinoTS}-LP, where only a lightweight linear forecasting head is trained. \textsc{WinoTS}-LP outperforms fully supervised TimeMixer on 7 datasets by MSE and 10 by MAE, demonstrating that pre-training extracts highly separable, informative temporal features without requiring full weight updates.

To assess whether these gains generalize beyond TimeMixer, Table~\ref{tab:backbone-relative-reduction} compares \textsc{WinoTS} against identical backbones trained from scratch under their default optimal supervised configurations. Across five datasets, \textsc{WinoTS} consistently reduces both MSE and MAE for all tested MLP, Transformer, and TCN-based backbones, confirming that its benefits are broadly model-agnostic. A detailed per-dataset breakdown for each backbone family is provided in Supplementary Table~\ref{tab:backbone-ablation}.

\begin{table}[tbh]
\centering
\caption{
Relative MSE/MAE reduction of \textsc{WinoTS} fine-tuning compared to training the same backbone architectures from scratch under their optimal supervised configurations. Results are averaged across five datasets ({ETTh1}, {ETTh2}, {ETTm1}, {ETTm2}, {Weather}) and four forecasting horizons ($H \in \{96, 192, 336, 720\}$). Positive values indicate performance gains.
}
\label{tab:backbone-relative-reduction}
\scriptsize
\begin{tabular}{llc}
\toprule
\textbf{Backbone Family} & \textbf{Encoder Backbone} & \textbf{Avg. MSE / MAE Reduction (\%)} \\
\midrule
\textbf{MLP} & TimeMixer & +3.2\% / +2.7\% \\
\midrule
\textbf{Transformer} & PatchTST & +3.5\% / +2.4\% \\
 & iTransformer & +2.6\% / +2.9\% \\
\midrule
\textbf{TCN} & TS2Vec & +58.0\% / +44.0\% \\
\bottomrule
\end{tabular}
\end{table}

\begin{table*}[h!]
\centering
\caption{
Cross-domain zero-shot forecasting transfer on ETT datasets (lower is better). Best and second-best results are highlighted in \textcolor{red}{red} and \textcolor{blue}{blue}, respectively. Each source$\rightarrow$target row denotes pre-training on the source dataset followed by direct evaluation on the target dataset, without target-domain parameter updates. 
}
\label{tab:cross-domain}
\scriptsize
\setlength{\tabcolsep}{2.3pt}
\resizebox{\textwidth}{!}{%
\begin{tabular}{lcccccccccccccccccccc}
\toprule
\textbf{Transfer} & \multicolumn{2}{c}{\textbf{\textsc{WinoTS}}} & \multicolumn{2}{c}{\textbf{\textsc{WinoTS}-LP}} & \multicolumn{2}{c}{\textbf{TimeMixer}} & \multicolumn{2}{c}{\textbf{TimeBase}} & \multicolumn{2}{c}{\textbf{SparseTSF}} & \multicolumn{2}{c}{\textbf{PatchTST}} & \multicolumn{2}{c}{\textbf{Autoformer}} & \multicolumn{2}{c}{\textbf{FEDformer}} & \multicolumn{2}{c}{\textbf{iTransformer}} & \multicolumn{2}{c}{\textbf{TimesNet}} \\
 & \multicolumn{2}{c}{(Ours)} & \multicolumn{2}{c}{(Ours)} & \multicolumn{2}{c}{\begin{tabular}[t]{@{}c@{}}\citeauthor{wang2024timemixer}\\ {[}\citeyear{wang2024timemixer}{]}\end{tabular}} & \multicolumn{2}{c}{\begin{tabular}[t]{@{}c@{}}\citeauthor{timebase}\\ {[}\citeyear{timebase}{]}\end{tabular}} & \multicolumn{2}{c}{\begin{tabular}[t]{@{}c@{}}\citeauthor{sparsetsf}\\ {[}\citeyear{sparsetsf}{]}\end{tabular}} & \multicolumn{2}{c}{\begin{tabular}[t]{@{}c@{}}\citeauthor{patch_tst}\\ {[}\citeyear{patch_tst}{]}\end{tabular}} & \multicolumn{2}{c}{\begin{tabular}[t]{@{}c@{}}\citeauthor{autoformer}\\ {[}\citeyear{autoformer}{]}\end{tabular}} & \multicolumn{2}{c}{\begin{tabular}[t]{@{}c@{}}\citeauthor{fedformer}\\ {[}\citeyear{fedformer}{]}\end{tabular}} & \multicolumn{2}{c}{\begin{tabular}[t]{@{}c@{}}\citeauthor{liu2023itransformer}\\ {[}\citeyear{liu2023itransformer}{]}\end{tabular}} & \multicolumn{2}{c}{\begin{tabular}[t]{@{}c@{}}\citeauthor{wu2023timesnet}\\ {[}\citeyear{wu2023timesnet}{]}\end{tabular}} \\
\cmidrule(lr){2-3} \cmidrule(lr){4-5} \cmidrule(lr){6-7} \cmidrule(lr){8-9} \cmidrule(lr){10-11} \cmidrule(lr){12-13} \cmidrule(lr){14-15} \cmidrule(lr){16-17} \cmidrule(lr){18-19} \cmidrule(lr){20-21}
 & MSE & MAE & MSE & MAE & MSE & MAE & MSE & MAE & MSE & MAE & MSE & MAE & MSE & MAE & MSE & MAE & MSE & MAE & MSE & MAE \\
\midrule
ETTh1$\rightarrow$ETTh2 & \textcolor{red}{0.3496} & \textcolor{red}{0.3888} & \textcolor{blue}{0.3520} & \textcolor{blue}{0.3902} & 0.3778 & 0.4000 & 0.3560 & 0.3991 & 0.3729 & 0.4075 & 0.3790 & 0.4020 & 0.4721 & 0.4810 & 0.4573 & 0.4696 & 0.3752 & 0.3991 & 0.4229 & 0.4307 \\
ETTh1$\rightarrow$ETTm1 & \textcolor{red}{0.7027} & \textcolor{blue}{0.5461} & \textcolor{blue}{0.7032} & \textcolor{red}{0.5457} & 0.7866 & 0.5738 & 0.7420 & 0.5647 & 0.8111 & 0.5644 & 0.8000 & 0.5890 & 0.7730 & 0.5870 & 0.7627 & 0.5802 & 0.8307 & 0.5851 & 0.9413 & 0.6222 \\
ETTh1$\rightarrow$ETTm2 & \textcolor{red}{0.2951} & \textcolor{red}{0.3481} & \textcolor{blue}{0.2955} & \textcolor{blue}{0.3482} & 0.3143 & 0.3559 & 0.3060 & 0.3611 & 0.3096 & 0.3616 & 0.3140 & 0.3570 & 0.3650 & 0.4070 & 0.3532 & 0.3902 & 0.3216 & 0.3632 & 0.3524 & 0.3837 \\
ETTh2$\rightarrow$ETTh1 & \textcolor{blue}{0.4473} & \textcolor{blue}{0.4512} & 0.4513 & 0.4536 & 0.6379 & 0.5457 & \textcolor{red}{0.4150} & \textcolor{red}{0.4156} & 0.5039 & 0.4739 & 0.6410 & 0.5490 & 0.7140 & 0.5820 & 0.6856 & 0.5760 & 0.6673 & 0.5676 & 0.8393 & 0.6436 \\
ETTh2$\rightarrow$ETTm1 & 0.7586 & \textcolor{red}{0.5666} & 0.7675 & \textcolor{blue}{0.5676} & 0.8627 & 0.5956 & 0.7620 & 0.5709 & 1.4956 & 0.6973 & 0.9680 & 0.6170 & \textcolor{blue}{0.7300} & 0.5730 & \textcolor{red}{0.7273} & 0.5720 & 0.9493 & 0.6235 & 1.3997 & 0.7322 \\
ETTh2$\rightarrow$ETTm2 & \textcolor{red}{0.2955} & \textcolor{red}{0.3495} & \textcolor{blue}{0.2956} & \textcolor{blue}{0.3504} & 0.3311 & 0.3708 & 0.3101 & 0.3649 & 0.3267 & 0.3759 & 0.3280 & 0.3680 & 0.3490 & 0.3840 & 0.3293 & 0.3718 & 0.3291 & 0.3689 & 0.3803 & 0.4030 \\
ETTm1$\rightarrow$ETTh1 & \textcolor{blue}{0.5392} & \textcolor{blue}{0.4945} & \textcolor{red}{0.5281} & \textcolor{red}{0.4919} & 0.7253 & 0.5786 & 0.6320 & 0.5417 & 0.5546 & 0.5066 & 0.6210 & 0.5390 & 0.9560 & 0.6590 & 1.057 & 0.704 & 0.7165 & 0.5676 & 1.0033 & 0.6814 \\
ETTm1$\rightarrow$ETTh2 & \textcolor{blue}{0.3908} & \textcolor{blue}{0.4208} & 0.3976 & 0.4236 & 0.4412 & 0.4407 & \textcolor{red}{0.3820} & \textcolor{red}{0.4191} & 0.4049 & 0.4257 & 0.4380 & 0.4380 & 0.4660 & 0.4730 & 0.4517 & 0.4605 & 0.4572 & 0.4499 & 0.5039 & 0.4820 \\
ETTm1$\rightarrow$ETTm2 & \textcolor{red}{0.2697} & \textcolor{red}{0.3216} & \textcolor{blue}{0.2724} & \textcolor{blue}{0.3228} & 0.2994 & 0.3353 & 0.2760 & 0.3302 & 0.2746 & 0.3264 & 0.2970 & 0.3340 & 0.3655 & 0.4072 & 0.323 & 0.366 & 0.2991 & 0.3341 & 0.3332 & 0.3648 \\
ETTm2$\rightarrow$ETTh1 & \textcolor{blue}{0.5011} & \textcolor{blue}{0.4842} & \textcolor{red}{0.4844} & \textcolor{red}{0.4733} & 0.7596 & 0.6059 & 0.7560 & 0.5918 & 0.6203 & 0.5419 & 0.6360 & 0.5590 & 0.7150 & 0.5770 & 1.265 & 0.743 & 0.8905 & 0.6348 & 1.0088 & 0.6678 \\
ETTm2$\rightarrow$ETTh2 & \textcolor{red}{0.3632} & \textcolor{red}{0.3972} & \textcolor{blue}{0.3660} & \textcolor{blue}{0.3988} & 0.4185 & 0.4322 & 0.3870 & 0.4197 & 0.3926 & 0.4138 & 0.4060 & 0.4210 & 0.4231 & 0.4363 & 0.429 & 0.445 & 0.4337 & 0.4441 & 0.4712 & 0.4595 \\
ETTm2$\rightarrow$ETTm1 & \textcolor{blue}{0.4791} & \textcolor{blue}{0.4552} & \textcolor{red}{0.4318} & \textcolor{red}{0.4301} & 0.5936 & 0.5067 & 0.5500 & 0.4880 & 1.0778 & 0.6149 & 0.6060 & 0.5110 & 0.7200 & 0.5653 & 0.720 & 0.565 & 0.6440 & 0.5193 & 0.8046 & 0.5813 \\
\bottomrule
\end{tabular}
}
\end{table*}
\subsection{Cross-Domain Zero-Shot Forecasting Transfer}
\label{sub:res_crossdomain}

To test whether \textsc{WinoTS} captures transferable temporal representations we further evaluate cross-domain forecasting transfer on the ETT benchmarks. In this setting, models are pre-trained on a source dataset and evaluated directly on an unseen target dataset without any target-domain fine-tuning.

As shown in Table~\ref{tab:cross-domain}, \textsc{WinoTS} exhibits particularly strong cross-domain transfer within the ETT benchmark family, 
achieving the best MSE on nine of the twelve source$\rightarrow$target pairs and the best MAE on ten, and the top average rank among all evaluated methods (1.42 by MSE and 1.17 by MAE). Relative to its direct {TimeMixer} base, \textsc{WinoTS} yields an average relative improvement of 16.4\% in MSE and 8.7\% in MAE, outperforming it across all transfer pairs.

\subsection{Unsupervised Anomaly Detection}
\label{sub:res_anomaly}
We further evaluate \textsc{WinoTS} representations on five standard multivariate anomaly-detection datasets under the TSLib evaluation protocol (Table~\ref{tab:anomaly-main}). Among the evaluated methods, \textsc{WinoTS} achieves the highest point-adjusted F1 on all five benchmarks.
These results indicate that the multi-scale representations learned by \textsc{WinoTS} transfer effectively to reconstruction-based anomaly detection. In particular, representations learned by matching detail-perturbed student views to denoised teacher targets appear well suited to modeling regular temporal structure and identifying deviations through reconstruction error.
\begin{table*}[t]
\centering
\caption{
Anomaly detection results.
We report point-adjusted precision (P), recall (R), and F1-score (\%). Best and second-best results are shown in
\textcolor{red}{red} and \textcolor{blue}{blue}, respectively.
}
\label{tab:anomaly-main}
\scriptsize
\setlength{\tabcolsep}{2.4pt}
\resizebox{\textwidth}{!}{%
\begin{tabular}{lcccccccccccccccccccccccc}
\toprule
Dataset & \multicolumn{3}{c}{\textsc{WinoTS}} & \multicolumn{3}{c}{iTransformer} & \multicolumn{3}{c}{DLinear} & \multicolumn{3}{c}{Autoformer} & \multicolumn{3}{c}{TimesNet} & \multicolumn{3}{c}{FEDformer} & \multicolumn{3}{c}{Crossformer} & \multicolumn{3}{c}{Reformer} \\
 & \multicolumn{3}{c}{(Ours)} & \multicolumn{3}{c}{\begin{tabular}[t]{@{}c@{}}\citeauthor{liu2023itransformer}\\ {[}\citeyear{liu2023itransformer}{]}\end{tabular}} & \multicolumn{3}{c}{\begin{tabular}[t]{@{}c@{}}\citeauthor{dlinear}\\ {[}\citeyear{dlinear}{]}\end{tabular}} & \multicolumn{3}{c}{\begin{tabular}[t]{@{}c@{}}\citeauthor{autoformer}\\ {[}\citeyear{autoformer}{]}\end{tabular}} & \multicolumn{3}{c}{\begin{tabular}[t]{@{}c@{}}\citeauthor{wu2023timesnet}\\ {[}\citeyear{wu2023timesnet}{]}\end{tabular}} & \multicolumn{3}{c}{\begin{tabular}[t]{@{}c@{}}\citeauthor{fedformer}\\ {[}\citeyear{fedformer}{]}\end{tabular}} & \multicolumn{3}{c}{\begin{tabular}[t]{@{}c@{}}\citeauthor{zhang2022crossformer}\\ {[}\citeyear{zhang2022crossformer}{]}\end{tabular}} & \multicolumn{3}{c}{\shortcite{reformer}}\\
\cmidrule(lr){2-4}\cmidrule(lr){5-7}\cmidrule(lr){8-10}\cmidrule(lr){11-13}\cmidrule(lr){14-16}\cmidrule(lr){17-19}\cmidrule(lr){20-22}\cmidrule(lr){23-25}
& P & R & F1 & P & R & F1 & P & R & F1 & P & R & F1 & P & R & F1 & P & R & F1 & P & R & F1 & P & R & F1 \\
\midrule
SMD & \textcolor{red}{{84.04}} & \textcolor{red}{{77.34}} & \textcolor{red}{{80.55}} & 68.22 & 63.79 & 65.93 & 70.13 & \textcolor{blue}{{69.34}} & \textcolor{blue}{{69.73}} & 67.91 & 41.63 & 51.62 & \textcolor{blue}{{79.28}} & 54.20 & 64.39 & 60.86 & 52.23 & 56.22 & 62.99 & 62.32 & 62.65 & 64.03 & 61.56 & 62.77 \\
MSL & \textcolor{red}{{88.47}} & \textcolor{red}{{69.39}} & \textcolor{red}{{77.78}} & 53.62 & 13.79 & 21.94 & 69.45 & 25.38 & 37.18 & \textcolor{blue}{{82.14}} & \textcolor{blue}{{41.25}} & \textcolor{blue}{{54.92}} & 62.38 & 19.03 & 29.17 & 81.72 & 39.34 & 53.11 & 77.20 & 27.22 & 40.25 & 78.22 & 35.51 & 48.85 \\
SMAP & \textcolor{red}{{92.36}} & \textcolor{red}{{64.56}} & \textcolor{red}{{76.00}} & 57.84 & 8.45 & 14.75 & 69.09 & 14.14 & 23.47 & 73.57 & 20.13 & 31.61 & 68.28 & 13.39 & 22.38 & 70.91 & 16.94 & 27.35 & 70.42 & 16.24 & 26.39 & \textcolor{blue}{{77.41}} & \textcolor{blue}{{22.56}} & \textcolor{blue}{{34.94}} \\
SWaT & \textcolor{red}{{34.72}} & \textcolor{red}{{8.59}} & \textcolor{red}{{13.78}} & 6.67 & 1.34 & 2.23 & 5.96 & 1.19 & 1.98 & 9.83 & 1.95 & 3.26 & 4.17 & 0.84 & 1.39 & 10.07 & 2.00 & 3.34 & \textcolor{blue}{{28.92}} & \textcolor{blue}{{7.37}} & \textcolor{blue}{\textbf{11.75}} & 9.76 & 1.93 & 3.23 \\
PSM & \textcolor{red}{{99.33}} & \textcolor{red}{{86.51}} & \textcolor{red}{{92.48}} & 93.38 & 30.87 & 46.40 & 96.79 & \textcolor{blue}{{42.52}} & \textcolor{blue}{{59.08}} & \textcolor{blue}{{98.65}} & 14.01 & 24.53 & 81.76 & 31.17 & 45.14 & 98.05 & 14.66 & 25.51 & 93.49 & 34.85 & 50.77 & 91.50 & 24.71 & 38.92 \\
\bottomrule
\end{tabular}%
}
\end{table*}
\section{Ablation Studies}
\label{sec:ablations}

In this section, we conduct systematic ablation experiments to isolate the impact of core framework design choices. Unless specified otherwise, all ablation models are pre-trained on the source dataset and evaluated downstream under the frozen linear probing (\textsc{WinoTS}-LP) protocol across four standard forecasting horizons ($H \in \{96, 192, 336, 720\}$).

\paragraph{Augmentation Strategy.} To examine the role of our wavelet-based view generation, we substitute our proposed wavelet augmentations with augmentations commonly used in vision self-distillation, including noise jitter, cropping, and their combinations (Table~\ref{tab:aug-ablation-vision}, visualized in Supplementary Figure~\ref{fig:wavelet-view-visualization}). Overall, preserving full temporal length and phase integrity yields consistently better linear probing performance.

Specifically, cropping combined with jitter (jitter+crop and gaussian+crop) leads to noticeable performance degradation across most benchmarks, particularly on ETTm1 (MSE $0.438$ vs. $0.364$) and Weather (MSE $0.303$ vs. $0.238$). While simple time-domain noise (jitter) preserves sequence length and achieves competitive results on specific streams, our wavelet-domain view generation provides superior performance on four out of six datasets, with notable margins on ETTm1 and Weather. This suggests that modulating coefficients in the wavelet domain provides an effective mechanism for augmenting temporal data while keeping overall sequence continuity and long-term structure intact.

\begin{table}[h!]
\centering
\caption{
{Wavelet versus vision-like augmentations.} We compare wavelet-based augmentations to jitter and cropping. Best results in \textcolor{red}{red}, second-best in \textcolor{blue}{blue}.
}
\label{tab:aug-ablation-vision}
\scriptsize
\setlength{\tabcolsep}{3.0pt}
\begin{tabular}{lcccccccc}
\toprule
 & \multicolumn{2}{c}{wavelets} & \multicolumn{2}{c}{jitter} & \multicolumn{2}{c}{jitter+crop} & \multicolumn{2}{c}{gaussian+crop} \\
\cmidrule(lr){2-3}\cmidrule(lr){4-5}\cmidrule(lr){6-7}\cmidrule(lr){8-9}
\textbf{Dataset} & MSE & MAE & MSE & MAE & MSE & MAE & MSE & MAE \\
\midrule
ETTh1       & \textcolor{red}{0.416} & \textcolor{red}{0.423} & \textcolor{blue}{0.417} & \textcolor{blue}{0.426} & 0.422 & 0.426 & 0.424 & 0.428 \\
ETTh2       & \textcolor{red}{0.347} & \textcolor{red}{0.385} & \textcolor{blue}{0.349} & \textcolor{blue}{0.387} & 0.349 & 0.388 & 0.360 & 0.392 \\
ETTm1       & \textcolor{red}{0.364} & \textcolor{red}{0.386} & \textcolor{blue}{0.397} & \textcolor{blue}{0.414} & 0.438 & 0.443 & 0.398 & 0.418 \\
ETTm2       & \textcolor{blue}{0.252} & \textcolor{blue}{0.308} & \textcolor{red}{0.247} & \textcolor{red}{0.308} & 0.253 & 0.312 & 0.274 & 0.328 \\
Weather     & \textcolor{red}{0.238} & \textcolor{red}{0.273} & \textcolor{blue}{0.242} & \textcolor{blue}{0.278} & 0.303 & 0.320 & 0.263 & 0.294 \\
Electricity & \textcolor{blue}{0.167} & \textcolor{blue}{0.260} & \textcolor{red}{0.165} & 0.257 & 0.165 & 0.257 & 0.165 & \textcolor{red}{0.256} \\
\bottomrule
\end{tabular}
\end{table}

\paragraph{Pre-training Objective.} We analyze the impact of the pre-training objective by contrasting the invariance-based joint-embedding distillation of \textsc{WinoTS} against a hybrid formulation as well as generative and reconstruction-based alternatives (Table~\ref{tab:objective-ablation2}). Specifically, we evaluate: (i) our invariance-based approach  (ii) a multi-task hybrid objective (Invariance-based + MAE), (iii) MAE, (iv) NTP, and (v) a joint-embedding predictive architecture (JEPA)~\cite{ijepa}.

The results demonstrate that incorporating a localized reconstruction target in (invariance-based+MAE) provides no systematic improvement over pure self-distillation and occasionally degrades downstream performance. Furthermore, purely generative and reconstruction-based objectives (MAE and NTP) trail behind the invariance-based objective of \textsc{WinoTS} on the majority of benchmarks. While NTP exhibits localized advantages on specific datasets (such as ETTm1 and Weather) where its autoregressive forecasting bias aligns closely with the downstream task, it fails to match the overall consistency of \textsc{WinoTS}. 
\begin{table}[h!]
\centering
\caption{
Ablation of \textsc{WinoTS}'s objective (Invariance-based self distillation).
In-domain forecasting MSE/MAE averaged over horizons $\{96,192,336,720\}$.
Lower is better; best in \textcolor{red}{red}, second-best in \textcolor{blue}{blue}.
}
\setlength{\tabcolsep}{3pt}
\label{tab:objective-ablation2}
\scriptsize
\begin{tabular}{lcccccccccc}
\toprule
\textbf{Dataset} & \multicolumn{2}{c}{\begin{tabular}[t]{@{}c@{}}\textbf{Invariance-}\\ \textbf{based}\\ \textbf{(Ours)}\end{tabular}} & \multicolumn{2}{c}{\begin{tabular}[t]{@{}c@{}}\textbf{Invariance-}\\ \textbf{based+MAE}\end{tabular}} & \multicolumn{2}{c}{\textbf{MAE}} & \multicolumn{2}{c}{\textbf{NTP}} & \multicolumn{2}{c}{\textbf{JEPA}} \\
\cmidrule(lr){2-3}\cmidrule(lr){4-5}\cmidrule(lr){6-7}\cmidrule(lr){8-9}\cmidrule(lr){10-11}
 & MSE & MAE & MSE & MAE & MSE & MAE & MSE & MAE & MSE & MAE \\
\midrule
ETTh1   & \textcolor{red}{0.416} & \textcolor{red}{0.423} & \textcolor{blue}{0.418} & \textcolor{blue}{0.424} & 0.465 & 0.465 & 0.428 & 0.435 & 0.442 & 0.447 \\
ETTh2   & \textcolor{red}{0.347} & \textcolor{red}{0.385} & \textcolor{blue}{0.355} & \textcolor{blue}{0.389} & 0.449 & 0.457 & 0.407 & 0.427 & 0.390 & 0.426 \\
ETTm1   & 0.364 & \textcolor{blue}{0.386} & 0.370 & 0.388 & \textcolor{blue}{0.349} & 0.387 & \textcolor{red}{0.346} & \textcolor{red}{0.381} & 0.375 & 0.393 \\
ETTm2   & \textcolor{red}{0.252} & \textcolor{red}{0.308} & \textcolor{blue}{0.258} & \textcolor{blue}{0.313} & 0.294 & 0.344 & 0.263 & 0.320 & 0.274 & 0.332 \\
Weather & 0.238 & 0.273 & 0.241 & 0.274 & 0.266 & 0.294 & \textcolor{red}{0.226} & \textcolor{red}{0.265} & \textcolor{blue}{0.229} & \textcolor{blue}{0.266} \\
\bottomrule
\end{tabular}
\end{table}

\section{Conclusion}
\label{sec:conclusion}

\textsc{WinoTS} demonstrates that wavelet-domain joint-embedding distillation offers a flexible, model-agnostic paradigm for self-supervised time-series learning, shifting the focus from point-wise reconstruction to multi-scale structural invariants. 

\textbf{Limitations \& Future Work.} While our global joint-embedding objective excels at capturing stable multi-scale patterns, it can be outperformed on highly volatile streams by specialized autoregressive priors (e.g., NTP) that explicitly optimize step-by-step local transitions. Future work will explore adaptive, learnable wavelet bases and extend joint-embedding distillation to scale-free, multi-modal temporal systems.

\bibliographystyle{plainnat}
\bibliography{aaai2026}

\newpage
\appendix
\appendix

\section*{Appendix}

\addcontentsline{toc}{section}{Appendix}

\renewcommand{\thesection}{\Alph{section}}

\setcounter{section}{0}

\setcounter{table}{0}

\renewcommand{\thetable}{A\arabic{table}}

\setcounter{figure}{0}

\renewcommand{\thefigure}{A\arabic{figure}}
This appendix provides additional methodological details, implementation specifications, and extended experimental results supporting the main paper. We first expand the WINO-TS methodology, including the teacher--student optimization, wavelet view construction, signal-processing interpretation, and architectural implementation details omitted from the main text. We then present extended experimental protocols, optimization and reproducibility settings, detailed per-horizon forecasting results, additional evaluations on synthetic pre-training and classification, and comprehensive ablation studies covering the pre-training objective, augmentation strategy, wavelet design choices, backbone generalization, view pairing, and robustness to random initialization. Together, these materials provide sufficient detail to reproduce and further analyze the proposed framework.

\section{Additional Method Details}
\label{app:method-details}

The main paper defines the WINO-TS wavelet views and self-distillation objective. This section records additional teacher-update, signal-processing, architectural, and implementation details that are useful for reproducing and interpreting the method.

\subsection{Teacher--Student Optimization and Cross-View Objective}
\label{app:teacher-student-details}

At the beginning of pre-training, the teacher parameters are initialized from the student parameters. Gradients are applied only to the student. After each student update, the teacher backbone and projection head are updated by exponential moving average (EMA):
\begin{equation}
\theta_t \leftarrow \lambda\theta_t + (1-\lambda)\theta_s,
\qquad
\phi_t \leftarrow \lambda\phi_t + (1-\lambda)\phi_s.
\end{equation}
The momentum coefficient follows a cosine schedule from $\lambda_0=0.9996$ to $1.0$ over pre-training.

The teacher center is an EMA of the batch mean of the teacher logits. For a batch of $B$ samples and $Q$ easy teacher views per sample, let
\begin{equation}
\overline{z}_t
=
\frac{1}{BQ}
\sum_{b=1}^{B}\sum_{q=1}^{Q} z_{t,b}^{q}.
\end{equation}
The center is updated as
\begin{equation}
c \leftarrow m_c c + (1-m_c)\overline{z}_t,
\qquad
m_c=0.9.
\end{equation}
The student temperature is fixed at $\tau_s=0.1$. The teacher temperature is scheduled from $0.06$ to its final value $\tau_t=0.04$ over the first five pre-training epochs.

During pre-training, WINO-TS attaches a three-layer DINO projection MLP to the backbone. The hidden layers have width $2048$ and use GELU activations, and the MLP terminates in a $256$-dimensional bottleneck. The bottleneck is $\ell_2$-normalized and passed through a weight-normalized linear output layer of dimension $K=1024$. The projection head is discarded after pre-training, and the EMA teacher backbone is transferred to the downstream task.

For an input window $X$, the wavelet view generator constructs $Q$
easy views
\begin{equation}
\mathcal{E}(X)
=
\left\{
X_q^{\mathrm{easy}}
\right\}_{q=1}^{Q}
\end{equation}
and $V$ hard views
\begin{equation}
\mathcal{H}(X)
=
\left\{
X_v^{\mathrm{hard}}
\right\}_{v=1}^{V}.
\end{equation}
The teacher processes only the easy views, whereas the student
processes both the easy and hard views. The same easy-view
realizations are therefore forwarded through both networks, while the
hard views are forwarded only through the student.

Let $P_{t,q}^{\mathrm{easy}}$ denote the teacher distribution for
easy view $q$, and let $P_{s,u}^{\mathrm{easy}}$ and
$P_{s,v}^{\mathrm{hard}}$ denote the corresponding student
distributions. The valid easy-teacher/easy-student pairs are
\begin{equation}
\mathcal{I}_{\mathrm{ee}}
=
\left\{
(q,u):
1\leq q,u\leq Q,\ q\neq u
\right\},
\end{equation}
and all easy-teacher/hard-student pairs are valid:
\begin{equation}
\mathcal{I}_{\mathrm{eh}}
=
\left\{
(q,v):
1\leq q\leq Q,\ 1\leq v\leq V
\right\}.
\end{equation}
The objective is
\begin{equation}
\begin{split}
\mathcal{L}_{\mathrm{WINO}}
=
\frac{1}{
|\mathcal{I}_{\mathrm{ee}}|
+
|\mathcal{I}_{\mathrm{eh}}|
}
\Bigg[
&
\sum_{(q,u)\in\mathcal{I}_{\mathrm{ee}}}
H\!\left(
\operatorname{sg}
\left[
P_{t,q}^{\mathrm{easy}}
\right],
P_{s,u}^{\mathrm{easy}}
\right)
\\
+
&
\sum_{(q,v)\in\mathcal{I}_{\mathrm{eh}}}
H\!\left(
\operatorname{sg}
\left[
P_{t,q}^{\mathrm{easy}}
\right],
P_{s,v}^{\mathrm{hard}}
\right)
\Bigg].
\end{split}
\end{equation}

Consequently, both student view types contribute directly to
optimization. Each easy student view is supervised by the teacher
outputs from the other easy-view realizations, while every hard
student view is supervised by all easy teacher outputs. Only the
teacher--student pair corresponding to the identical easy-view index
is excluded.

\subsection{Multiresolution Interpretation}
\label{app:wavelet-mra}

For an orthonormal wavelet basis, a $J$-level discrete wavelet transform decomposes a single-channel signal into an approximation space $V_J$ and its multi-scale orthogonal complement:
\begin{equation}
x
=
\underbrace{\sum_k a_{J,k}\phi_{J,k}}_{P_{V_J}x}
+
\underbrace{\sum_{j=1}^{J}\sum_k d_{j,k}\psi_{j,k}}_{\in V_J^\perp}.
\end{equation}
The approximation coefficients encode coarse temporal structure, whereas $d_{j,k}$ describes localized variation at a temporal scale on the order of $2^j$ around location $2^j k$. Unlike Fourier atoms, which have global temporal support, wavelet atoms are localized in both time and scale. This makes the representation suitable for non-stationary signals whose local frequency content changes over time.

A wavelet with $N$ vanishing moments annihilates polynomials of degree below $N$:
\begin{equation}
\psi \perp \{1,t,\ldots,t^{N-1}\}.
\end{equation}
Equivalently, the corresponding scaling filter contains an $N$-fold zero at $\omega=\pi$:
\begin{align}
H(\omega)
&=
\left(
\frac{1+e^{-i\omega}}{2}
\right)^N
Q(e^{-i\omega}),
\\
H(\omega)
&=
\frac{1}{\sqrt{2}}
\sum_k h_k e^{-ik\omega}.
\end{align}
with the orthonormal perfect-reconstruction condition
\begin{equation}
|H(\omega)|^2 + |H(\omega+\pi)|^2 = 1.
\end{equation}
The number of vanishing moments therefore controls how strongly low-order trends are suppressed by the detail filters and affects the sparsity of wavelet coefficients on smooth signals.

The critically sampled DWT is not shift-invariant: a small temporal shift can alter coefficient alignment across levels. WINO-TS treats this basis sensitivity as an additional source of view diversity rather than attempting to remove it. At the default depth $J=3$, the approximation emphasizes temporal structure at scales longer than roughly $2^{J+1}$ samples, while the detail bands retain progressively finer local fluctuations.

\paragraph{Why full-length temporal views.}
DINO-style learning depends strongly on the augmentation family. Image transformations such as multi-crop, color jitter, and blur generally alter nuisance factors while retaining object identity, but they do not have direct semantic analogues for ordered temporal signals. Random cropping can remove the context needed to identify trend, seasonality, or regime; time warping can distort phase relationships; and unrestricted amplitude jitter or additive noise can erase discriminative local structure. WINO-TS therefore realizes view diversity through localized wavelet coefficients while reconstructing every view at the original length.

\begin{figure*}[t]
    \centering
    \includegraphics[width=\textwidth]{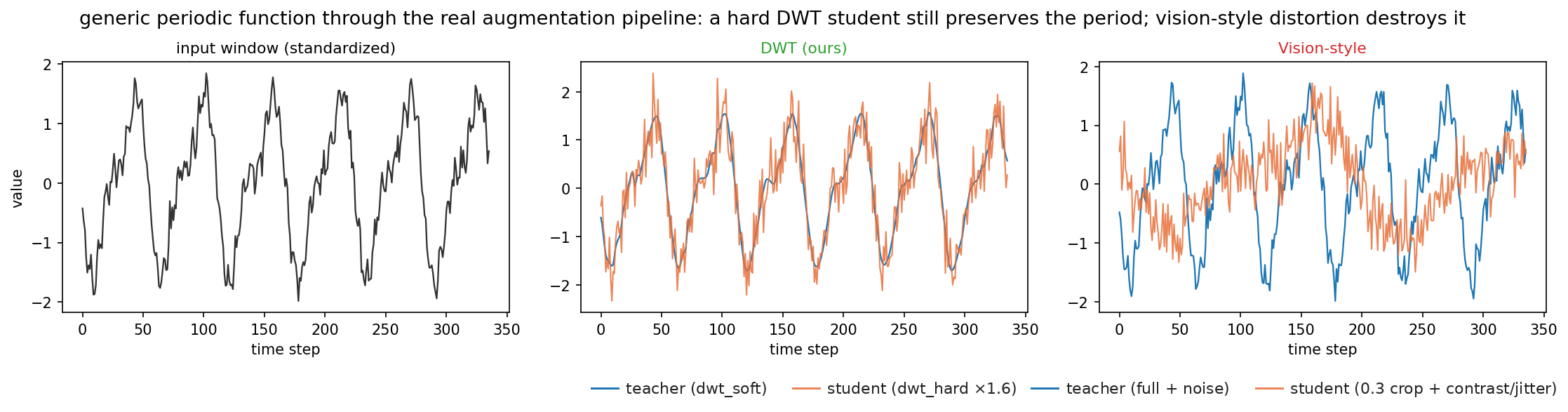}
    \caption{
    Qualitative comparison of wavelet-domain and vision-style view
    generation on a standardized periodic input window.
    The left panel shows the original input signal.
    In the middle panel, the WINO-TS easy teacher view, produced through
    wavelet-detail soft thresholding, and the hard student view, produced
    through wavelet-detail perturbation, retain the dominant period,
    temporal ordering, and full sequence support. Their differences are
    concentrated primarily in localized amplitude and high-frequency
    fluctuations.
    In the right panel, a representative vision-style augmentation
    pipeline based on temporal cropping, contrast modification, jitter,
    and additive noise alters the support and point-wise alignment of the
    two views, making their periodic correspondence less direct.
    The example illustrates the intended inductive bias of WINO-TS:
    preserving coarse temporal organization and global context while
    varying localized fine-scale content. It is provided as a qualitative
    sanity check rather than as general evidence of semantic preservation.
    }
    \label{fig:wavelet-view-visualization}
\end{figure*}

\paragraph{Qualitative visualization of temporal-structure preservation.}
Figure~\ref{fig:wavelet-view-visualization} illustrates the difference
between the proposed wavelet-domain view generator and a representative
vision-style augmentation pipeline. Under WINO-TS, the easy teacher and
hard student views remain temporally aligned and preserve the dominant
periodic structure of the original window, despite differing in their
localized high-frequency content. This produces a non-trivial
self-distillation task without removing temporal context or changing the
sequence support. In contrast, cropping and unstructured time-domain
perturbations can alter support, alignment, and phase correspondence
between teacher and student views. The visualization therefore
illustrates why full-length, frequency-localized perturbations provide a
more natural invariance mechanism for periodic temporal signals.

\subsection{Easy View and Its Relation to Classical Wavelet Shrinkage}
\label{app:easy-view-shrinkage}

For a selected wavelet basis, the easy view preserves the approximation coefficients and applies soft-threshold shrinkage independently to each detail band:
\begin{equation}
x^{\mathrm{easy}}
=
\mathcal{W}_{w}^{-1}
\left(
a_J,
\eta_{\tau_J}(d_J),
\ldots,
\eta_{\tau_1}(d_1)
\right),
\end{equation}
where
\begin{equation}
\eta_{\tau}(d)
=
\operatorname{sign}(d)\max(|d|-\tau,0),
\qquad
\tau_j
=
\rho\max_k |d_{j,k}|.
\end{equation}
Equivalently,
\begin{equation}
x^{\mathrm{easy}}
=
P_{V_J}x
+
\sum_{j,k}
\eta_{\tau_j}(d_{j,k})\psi_{j,k}.
\end{equation}
The easy view is therefore not a pure low-pass projection: it retains attenuated versions of dominant detail coefficients while suppressing weak transients.

This operation is inspired by classical wavelet shrinkage \cite{donoho_johnstone}, but it does not use the Donoho--Johnstone noise-calibrated threshold. WINO-TS instead uses a per-band threshold scaled by the largest absolute coefficient in that band. This avoids estimating a separate observation-noise model for every dataset and adapts the threshold to the dynamic range of each level.

Soft-thresholding is nonlinear and generally not idempotent:
\begin{equation}
\eta_\tau(\eta_\tau(d))
\neq
\eta_\tau(d).
\end{equation}
Accordingly, it does not project directly onto $V_J$. The default $\rho=0.6$ applies strong but partial shrinkage. When $\rho\geq 1$, every detail coefficient is mapped to zero and the construction reaches the hard low-pass limit $P_{V_J}x$.

\subsection{Hard View and Standardized Perturbation Scale}
\label{app:hard-view-details}

The hard view preserves the approximation coefficients and injects Gaussian perturbations into all detail levels:
\begin{align}
x^{\mathrm{hard}}
&=
\mathcal{W}_{w}^{-1}
\left(
a_J,
d_J+\varepsilon_J,
\ldots,
d_1+\varepsilon_1
\right),
\\
\end{align}
\begin{equation}
s_{v,c}
\sim
\mathcal{U}(0.2,0.5),
\qquad
\varepsilon_{v,j,k,c}
\sim
\mathcal{N}(0,s_{v,c}^{2}).
\end{equation}
\hl{For each hard view $v$, a scale $s_{v,c}$ is sampled independently
for every channel $c$ and shared across all detail levels of that
channel.} The construction can also be written as
\begin{equation}
x^{\mathrm{hard}}
=
x+\mathcal{W}_{w}^{-1}M_{\mathrm{HP}}\varepsilon,
\end{equation}
where $M_{\mathrm{HP}}$ selects the high-pass detail coefficients.

Inputs are standardized channel-wise using statistics computed only from the training split before the wavelet transform is applied. The perturbation scale $s$ is therefore expressed in standardized signal units. For multivariate inputs, the same wavelet basis is shared across channels within a view, but each channel receives an independent realization of $\varepsilon$. This preserves a coordinated view-level transform without imposing artificial high-frequency co-dependence between variables.

\subsection{Boundary Handling, Full-Length Reconstruction, and Admissible Depth}
\label{app:boundary-depth}

Finite windows require an extension rule at their boundaries. WINO-TS uses symmetric boundary extension. The ideal decomposition into $P_{V_J}x$ and $V_J^\perp$ therefore holds exactly in the interior, with boundary-dependent coefficients confined to a region whose order scales as $\mathcal{O}(2^J L)$ per side for a filter of length $L$. This is an order statement rather than an exact count, because the affected region depends on the implementation and wavelet support.

When the same basis is used for two views, their unmodified approximation coefficients agree up to the boundary convention. Under the default independent basis sampling, the easy and hard views instead preserve comparable low-frequency bands; their approximation coefficients need not be identical.

The decomposition depth must be compatible with the input length and filter support. The admissible depth satisfies
\begin{equation}
J
\leq
\left\lfloor
\log_2
\left(
\frac{T}{L-1}
\right)
\right\rfloor.
\end{equation}
For the longest filter in the default pool, $\mathrm{sym}8$ with $L=16$, the default depth $J=3$ requires
\begin{equation}
T
\geq
2^3(L-1)
=
120.
\end{equation}
All views are reconstructed to the original length $T$, so the augmentation does not require changes to a backbone's positional encoding, patch layout, or input-shape parameters.

\subsection{Wavelet Pool and Basis Diversity}
\label{app:wavelet-pool-details}

\begin{gather}
w^{\mathrm{easy}},\, w^{\mathrm{hard}}
\overset{\mathrm{i.i.d.}}{\sim}
\mathcal{U}(\mathcal{P}), \\
\mathcal{P}
=
\{\mathrm{sym}4,\, \mathrm{sym}6,\, \mathrm{sym}8,\, \mathrm{db}4,\, \mathrm{db}6,\, \mathrm{coif}2\}.
\end{gather}

\begin{table}[t]
\centering
\caption{
The WINO-TS wavelet pool $\mathcal{P}$.
VM denotes the number of vanishing moments of the wavelet function $\psi$, and $L$ denotes filter length in samples.
}
\label{tab:app-wavelet-pool}
\scriptsize
\begin{tabular}{@{}llccl@{}}
\toprule
\textbf{Wavelet} & \textbf{Family} & \textbf{VM} & $\boldsymbol{L}$ & \textbf{Primary property} \\
\midrule
sym4  & Symlet      & 4 & 8  & Reduced phase asymmetry \\
sym6  & Symlet      & 6 & 12 & Reduced phase asymmetry \\
sym8  & Symlet      & 8 & 16 & Reduced phase asymmetry \\
db4   & Daubechies  & 4 & 8  & Minimum-phase design \\
db6   & Daubechies  & 6 & 12 & Minimum-phase design \\
coif2 & Coiflet     & 4 & 12 & Moment constraints on $\phi$ and $\psi$ \\
\bottomrule
\end{tabular}
\end{table}

The pool varies support length, smoothness, vanishing moments, and phase behavior. Daubechies wavelets are compactly supported and minimum phase; Symlets reduce phase asymmetry; and Coiflets impose moment constraints on both the scaling and wavelet functions. Independent sampling prevents the encoder from binding its representation to one basis and introduces mild cross-basis phase and boundary variation while retaining comparable coarse temporal content.

\subsection{Model-Agnostic Transfer}
\label{app:model-agnostic-transfer}

At the objective level, WINO-TS requires only a backbone that maps a full-length time-series window to a representation accepted by the projection head. The backbone may be an MLP, Transformer, or temporal convolutional network. After pre-training, the projection head is removed and the EMA teacher backbone is used for downstream adaptation.

Two forecasting adaptation modes are considered. In linear probing, the backbone is frozen and only a task-specific forecasting head is optimized. In full fine-tuning, the backbone and task head are optimized jointly. Classification uses full fine-tuning: the pretrained backbone is unfrozen and optimized jointly with the classification head. In zero-shot cross-domain forecasting, pre-training and task adaptation are performed only on the source dataset, followed by direct target evaluation without target-domain parameter updates.

\section{Additional Positioning and Related Work}
\label{app:additional-positioning}

WINO-TS is a pre-training and view-construction method rather than a new supervised forecasting architecture. It is therefore complementary to backbones such as PatchTST, iTransformer, and TimeMixer: the DINO projection head is attached only during pre-training and is removed before downstream adaptation.

A related controlled study examined the pre-training dividend in
time-series foundation models \cite{major2026quantifying}. It compared
generative objectives, including masked reconstruction, next-token
prediction, and diffusion, with latent-alignment objectives, including
JEPA, Le-JEPA, and DINO, under a shared evaluation protocol, and
identified a task-dependent precision--invariance trade-off. WINO-TS
addresses a different question: it develops the DINO direction into a
dedicated time-series pre-training recipe based on wavelet-domain view
construction, basis-family sampling, and full-length multivariate
adaptation, and evaluates the resulting representations across
forecasting, transfer, anomaly detection, and classification.

Recent time-series self-distillation methods also use non-contrastive teacher--student learning, but differ in how they define the prediction task. TimeSiam uses past/current subseries and masked past-to-current reconstruction. Self-Distilled Representation Learning for Time Series follows a data2vec-style latent-prediction objective from masked inputs. HiMTM combines hierarchical masked time-series modeling with self-distillation, and UTICA adapts DINOv2-style training to time-series classification. WINO-TS instead constructs full-length views in the wavelet domain and varies localized detail coefficients through shrinkage, perturbation, and basis sampling rather than cropping, subseries selection, or reconstruction-only targets.

\section{Extended Experimental Protocols}
\label{app:experimental-protocols}

\subsection{Pre-Training and Adaptation}
\label{app:pretraining-adaptation}

For in-domain forecasting, the backbone is first pre-trained on the unlabeled training split of the same dataset. The projection head is then discarded and the EMA teacher backbone is adapted using either full fine-tuning or linear probing.

For cross-domain zero-shot forecasting, pre-training and task adaptation are performed strictly on the source dataset. The resulting source model is evaluated directly on the target dataset without target-domain fine-tuning or other target-domain parameter updates.

For classification, a classification head is attached to the
WINO-TS-pretrained backbone, and the complete model is fine-tuned
end-to-end with supervision. Both the pretrained backbone and the
classification head are updated. Forecasting results use MSE and MAE, averaged over horizons $\{96,192,336,720\}$ unless a table states otherwise. Classification results use accuracy.

\subsection{Dataset Naming} \label{app:dataset-naming}

The main paper reports results on the standard long-term forecasting benchmarks ETTh1, ETTh2, ETTm1, ETTm2, Weather, Electricity, Exchange, Solar, and Traffic. The appendix adds further datasets for a broader robustness evaluation.

For readability, we use the following short names throughout: AirQuality-Shunyi and AirQuality-Wan (also written AQShunyi and AQWan), PM2.5, JapaneseVowels, and SelfRegulationSCP1/2. ``Solar Energy'' is abbreviated to Solar; CzeLan keeps the capitalization of the main tables. All other forecasting names match the main experimental tables.

\section[Optimization and Reproducibility]{\raggedright Optimization and Reproducibility}
\label{app:optim}

For completeness, we report the optimization and compute configuration used for pre-training and downstream evaluation. All runs use seed~42 unless a table explicitly reports multiple seeds.

\subsection{Self-Supervised Pre-Training}
\label{app:optim-pretraining}

The WINO-TS self-distillation objective is optimized with AdamW. The base learning rate is $5\times10^{-4}$ and is scaled linearly with the effective batch size,
\begin{equation}
\eta_{\mathrm{eff}}
=
\eta_{\mathrm{base}}
\frac{B N_{\mathrm{GPU}}}{256}.
\end{equation}
The learning rate is warmed up linearly over the first three epochs and then decayed to $1\times10^{-6}$ with a cosine schedule. Weight decay follows a cosine schedule from $0.04$ to $0.1$, and gradients are clipped to a maximum norm of $3.0$. The EMA teacher momentum is cosine-annealed from $0.9996$ to $1.0$. The teacher temperature is scheduled from $0.06$ to $0.04$ over the first five epochs, while the student temperature remains fixed at $0.1$. The final prototype layer is frozen during the first epoch to stabilize early training.

Pre-training runs for 80 epochs in full precision (FP32) with a per-GPU batch size of 128. Each input contains 336 time steps, and the projection head outputs $K=1024$ prototypes.

\begin{table}[t]
\centering
\caption{WINO-TS pre-training hyperparameters.}
\label{tab:app-pretrain-hparams}
\scriptsize
\setlength{\tabcolsep}{5pt}
\begin{tabular}{ll}
\toprule
\textbf{Hyperparameter} & \textbf{Value} \\
\midrule
Optimizer                 & AdamW \\
Base learning rate        & $5\times10^{-4}$ (linear batch scaling, $/256$) \\
LR schedule               & $3$-epoch warmup $\to$ cosine to $1\times10^{-6}$ \\
Weight decay              & cosine $0.04 \to 0.1$ \\
Gradient clipping         & max-norm $3.0$ \\
Epochs                    & $80$ \\
Batch size (per GPU)      & $128$ \\
Precision                 & FP32 \\
EMA teacher momentum      & cosine $0.9996 \to 1.0$ \\
Teacher temperature       & warmup $0.06 \to 0.04$ ($5$ epochs) \\
Student temperature       & $0.1$ \\
Freeze last layer         & $1$ epoch \\
Easy (teacher) views      & $1$ \\
Hard (student) views      & $1$ \\
Input window / patch      & $336$ \\
Prototypes ($K$)          & $1024$ \\
Backbone                  & TimeMixer, $4$ layers, $d_{\mathrm{model}}{=}128$,\\
                          & $d_{\mathrm{ff}}{=}256$, dropout $0.1$ \\
Attention pooling         & multi-head attention, $d_{\mathrm{model}}{=}128$,\\
                          & $4$ heads, dropout $0.0$ \\
\bottomrule
\end{tabular}
\end{table}

\subsection{Forecasting}
\label{app:optim-forecasting}

Downstream forecasting uses Adam with a one-cycle learning-rate schedule comprising a 30\% warm-up phase followed by cosine decay, together with weight decay $1\times10^{-4}$. We report two adaptation protocols. In linear probing, the backbone is frozen and only the forecasting head is trained. In full fine-tuning, the backbone is unfrozen and the encoder and head use a shared learning rate. The learning rate is $2\times10^{-5}$ for linear probing and $1\times10^{-4}$ for full fine-tuning. A subset of fine-tuning runs instead uses the learning-rate-free Prodigy~\citep{mishchenko2024prodigy} optimizer with $d_{\mathrm{coef}}\in\{0.5,1.0\}$.

Fine-tuning runs for approximately 20 epochs with early stopping on a held-out validation split; the checkpoint with the best validation performance is retained. The default forecasting batch size is 128 and is reduced for the highest-channel
datasets to fit in memory: 32 for PM2.5, 16 for Electricity, and 8 for Traffic. Head dropout is selected from $\{0.0,0.1\}$. The same optimization configuration is used for horizons $\{96,192,336,720\}$.

\subsection{Anomaly Detection}
\label{app:optim-anomaly}

\hl{The anomaly detector consists of a WINO-TS-pretrained encoder followed by a linear
reconstruction decoder. The encoder and decoder are jointly fine-tuned, using normal windows from the training split.}
The detector
is optimized with Adam for 10 epochs using a learning rate
of $1\times10^{-3}$ and a mean-squared reconstruction loss over
non-overlapping windows of length 100, represented as 10 patches of
length 10.

Given an input $X$ and its reconstruction $\widehat{X}$, the anomaly
score at timestamp $t$ is the channel-averaged squared reconstruction
error:
\begin{equation}
e_t
=
\frac{1}{C}
\left\|
X_t-\widehat{X}_t
\right\|_2^2.
\end{equation}

Following the TSLib evaluation protocol, the decision threshold is
computed from the combined training and unlabeled test reconstruction
scores:
\begin{equation}
\gamma_d
=
Q_{1-r_d/100}
\left(
\mathcal{E}^{d}_{\mathrm{train}}
\cup
\mathcal{E}^{d}_{\mathrm{test}}
\right),
\end{equation}
where $\mathcal{E}^{d}_{\mathrm{train}}$ and
$\mathcal{E}^{d}_{\mathrm{test}}$ denote the timestamp-level
reconstruction-energy distributions for dataset $d$. We use
$r_{\mathrm{SMD}}=0.5$ and $r_d=1.0$ for MSL, SMAP, SWaT,
and PSM, corresponding to the $99.5$th and $99$th percentiles,
respectively. These dataset-specific anomaly ratios are predefined
by the evaluation configuration and are not estimated from the
empirical test anomaly rate. Test labels are not used to determine
the threshold.

A test timestamp is initially classified as anomalous when
$e_t^{\mathrm{test}}>\gamma_d$. We then apply the standard
segment-level point-adjustment procedure before computing precision,
recall, and F1. All methods in the anomaly comparison use the same
scoring, thresholding, and point-adjustment procedure.

\subsection{Classification}
\label{app:optim-classification}
Classification uses full end-to-end fine-tuning. A linear
classification head is attached to the pretrained backbone, and both
the backbone and classification head are jointly optimized with
Adam at learning rate $1\times10^{-3}$ for 20 epochs using
batch size 16 on fixed-length windows.

\subsection{Hardware and Software}
\label{app:optim-hardware}

All experiments run on a server with $8\times$ NVIDIA RTX PRO 6000 Blackwell Server Edition GPUs with 96~GB of memory per GPU. Each dataset--model run uses one GPU; no multi-GPU training is used. The software stack is Python~3.9.25 and PyTorch~2.8.0 with CUDA~12.8 and cuDNN~9.10.

Pre-training cost scales with the number of channels. On one GPU, an epoch takes approximately 24~s for ETTh1 (7 channels) and approximately 26~min for Electricity (321 channels); a complete ETT pre-training run takes approximately 32~min. Downstream fine-tuning takes approximately 10--15~min on smaller datasets and up to a few hours on the largest datasets.

\begin{table}[t]
\centering
\caption{
Forecasting (MSE/MAE, avg. over horizons $\{96,192,336,720\}$) with \emph{synthetic} pre-training.
Ours (FT) = best fine-tuned DINO config per dataset; Ours (LP) and survey \cite{major2026quantifying} are linear-probe.
Each MSE/MAE pair is from a single run. Lower is better.
}
\label{tab:synth-prev-vs-now}
\scriptsize
\setlength{\tabcolsep}{4pt}
\begin{tabular}{lcccccc}
\toprule
& \multicolumn{2}{c}{\begin{tabular}[t]{@{}c@{}}\textbf{Ours}\\ \textbf{(synth, FT)}\end{tabular}} 
& \multicolumn{2}{c}{\begin{tabular}[t]{@{}c@{}}\textbf{Ours}\\ \textbf{(synth, LP)}\end{tabular}} 
& \multicolumn{2}{c}{\begin{tabular}[t]{@{}c@{}}\textbf{Survey}\\ \textbf{(synth, LP)}\end{tabular}} \\
\cmidrule(lr){2-3}\cmidrule(lr){4-5}\cmidrule(lr){6-7}
\textbf{Dataset} & MSE & MAE & MSE & MAE & MSE & MAE \\
\midrule
ETTh1   & 0.417 & 0.425 & 0.521 & 0.489 & 0.438 & 0.446 \\
ETTh2   & 0.365 & 0.402 & 0.410 & 0.428 & 0.362 & 0.401 \\
ETTm1   & 0.351 & 0.378 & 0.361 & 0.389 & 0.357 & 0.384 \\
ETTm2   & 0.250 & 0.310 & 0.259 & 0.319 & 0.253 & 0.311 \\
Weather & 0.227 & 0.261 & 0.242 & 0.276 & 0.235 & 0.272 \\
\bottomrule
\end{tabular}
\end{table}

\begin{table}[h!]
\centering
\caption{Effect of the DINO head output dimension $K$ (out\_dim) on in-domain
forecasting (linear probe, context 336, MSE/MAE averaged over horizons
$\{96,192,336,720\}$). Ours uses $K{=}1024$; best per row in \textbf{bold}.}
\label{tab:outdim-ablation}
\scriptsize
\setlength{\tabcolsep}{3pt}
\begin{tabular}{lcccccccc}
\toprule
 & \multicolumn{2}{c}{$K{=}512$} & \multicolumn{2}{c}{\begin{tabular}[t]{@{}c@{}}$K{=}1024$\\ (Ours)\end{tabular}} & \multicolumn{2}{c}{$K{=}2048$} & \multicolumn{2}{c}{$K{=}8192$} \\
\cmidrule(lr){2-3}\cmidrule(lr){4-5}\cmidrule(lr){6-7}\cmidrule(lr){8-9}
Dataset & MSE & MAE & MSE & MAE & MSE & MAE & MSE & MAE \\
\midrule
ETTh1   & 0.430 & 0.431 & \textbf{0.416} & \textbf{0.423} & 0.431 & 0.433 & 0.435 & 0.434 \\
ETTh2   & 0.369 & 0.395 & \textbf{0.347} & \textbf{0.385} & 0.377 & 0.399 & 0.378 & 0.400 \\
ETTm1   & 0.364 & 0.387 & 0.364 & 0.386 & 0.359 & \textbf{0.383} & \textbf{0.357} & 0.384 \\
ETTm2   & \textbf{0.248} & 0.311 & 0.252 & \textbf{0.308} & \textbf{0.248} & 0.310 & 0.249 & 0.309 \\
Weather & 0.238 & 0.273    & \textbf{0.238} & \textbf{0.273} & 0.269    & 0.298 & 0.241 & 0.275    \\
\bottomrule
\end{tabular}
\end{table}

\section{Extended Forecasting Results}
\label{app:extended-forecasting}

\subsection{Complete In-Domain Averages}
\label{app:full-forecasting-reference}

The complete in-domain forecasting comparison over all 13 datasets is already reported in the main paper in Table~\ref{tab:forecasting-main}. That table includes WINO-TS under full fine-tuning and linear probing, the self-supervised TimeSiam and TS2Vec baselines, and all supervised forecasting baselines, with MSE and MAE averaged over horizons $\{96,192,336,720\}$. We refer to the main-paper table rather than reproducing the same large table in the appendix.

\subsection{Horizon-wise Forecasting Results}
\label{app:horizon-wise}
Table~\ref{tab:forecasting-perhorizon} provides the complete per-horizon breakdown for all 13 in-domain forecasting benchmarks. It reports horizons $\{96,192,336,720\}$ under the same input context length of 336 and includes both WINO-TS adaptation protocols, the self-supervised baselines, and all supervised baselines used in the main comparison.

\begin{table*}[h!]
\centering
\caption{
In-domain forecasting performance per prediction length $\{96,192,336,720\}$ (lower is better).
All methods use the same input context length of 336.
Best and second-best per row are shown in \textcolor{red}{red} and \textcolor{blue}{blue}.
\textbf{Ours}: WINO-TS (full fine-tune) and WINO-TS-LP (frozen backbone, trained head).
\textbf{Self-supervised}: TimeSiam and TS2Vec. \textbf{Supervised}: end-to-end baselines.
}
\label{tab:forecasting-perhorizon}
\scriptsize
\setlength{\tabcolsep}{2.0pt}
\resizebox{\textwidth}{!}{%
\begin{tabular}{llcccccccccccccccccccccccccc}
\toprule
 & & \multicolumn{4}{c}{\textbf{Ours}} & \multicolumn{4}{c}{\textbf{Self-Supervised}} & \multicolumn{18}{c}{\textbf{Supervised}} \\
\cmidrule(lr){3-6}\cmidrule(lr){7-10}\cmidrule(lr){11-28}
\textbf{Dataset} & \textbf{H} & \multicolumn{2}{c}{WINO-TS} & \multicolumn{2}{c}{\begin{tabular}[t]{@{}c@{}}WINO-\\ TS-LP\end{tabular}} & \multicolumn{2}{c}{TimeSiam} & \multicolumn{2}{c}{TS2Vec} & \multicolumn{2}{c}{TimeMixer} & \multicolumn{2}{c}{TimeBase} & \multicolumn{2}{c}{SparseTSF} & \multicolumn{2}{c}{PatchTST} & \multicolumn{2}{c}{DLinear} & \multicolumn{2}{c}{iTransformer} & \multicolumn{2}{c}{FEDformer} & \multicolumn{2}{c}{TimesNet} & \multicolumn{2}{c}{Autoformer} \\
 & & \multicolumn{2}{c}{(Ours)} & \multicolumn{2}{c}{(Ours)} & \multicolumn{2}{c}{\begin{tabular}[t]{@{}c@{}}\citeauthor{timesiam}\\ {[}\citeyear{timesiam}{]}\end{tabular}} & \multicolumn{2}{c}{\begin{tabular}[t]{@{}c@{}}\citeauthor{ts2vec}\\ {[}\citeyear{ts2vec}{]}\end{tabular}} & \multicolumn{2}{c}{\begin{tabular}[t]{@{}c@{}}\citeauthor{wang2024timemixer}\\ {[}\citeyear{wang2024timemixer}{]}\end{tabular}} & \multicolumn{2}{c}{\begin{tabular}[t]{@{}c@{}}\citeauthor{timebase}\\ {[}\citeyear{timebase}{]}\end{tabular}} & \multicolumn{2}{c}{\begin{tabular}[t]{@{}c@{}}\citeauthor{sparsetsf}\\ {[}\citeyear{sparsetsf}{]}\end{tabular}} & \multicolumn{2}{c}{\begin{tabular}[t]{@{}c@{}}\citeauthor{patch_tst}\\ {[}\citeyear{patch_tst}{]}\end{tabular}} & \multicolumn{2}{c}{\begin{tabular}[t]{@{}c@{}}\citeauthor{dlinear}\\ {[}\citeyear{dlinear}{]}\end{tabular}} & \multicolumn{2}{c}{\begin{tabular}[t]{@{}c@{}}\citeauthor{liu2023itransformer}\\ {[}\citeyear{liu2023itransformer}{]}\end{tabular}} & \multicolumn{2}{c}{\begin{tabular}[t]{@{}c@{}}\citeauthor{fedformer}\\ {[}\citeyear{fedformer}{]}\end{tabular}} & \multicolumn{2}{c}{\begin{tabular}[t]{@{}c@{}}\citeauthor{wu2023timesnet}\\ {[}\citeyear{wu2023timesnet}{]}\end{tabular}} & \multicolumn{2}{c}{\begin{tabular}[t]{@{}c@{}}\citeauthor{autoformer}\\ {[}\citeyear{autoformer}{]}\end{tabular}} \\
\cmidrule(lr){3-4}\cmidrule(lr){5-6}\cmidrule(lr){7-8}\cmidrule(lr){9-10}\cmidrule(lr){11-12}\cmidrule(lr){13-14}\cmidrule(lr){15-16}\cmidrule(lr){17-18}\cmidrule(lr){19-20}\cmidrule(lr){21-22}\cmidrule(lr){23-24}\cmidrule(lr){25-26}\cmidrule(lr){27-28}
 & & MSE & MAE & MSE & MAE & MSE & MAE & MSE & MAE & MSE & MAE & MSE & MAE & MSE & MAE & MSE & MAE & MSE & MAE & MSE & MAE & MSE & MAE & MSE & MAE & MSE & MAE \\
\midrule
\multirow{4}{*}{ETTh1} & 96 & \textcolor{red}{0.369} & \textcolor{blue}{0.394} & 0.374 & 0.396 & 0.376 & 0.405 & 0.647 & 0.578 & 0.379 & 0.402 & \textcolor{blue}{0.370} & \textcolor{red}{0.385} & 0.374 & \textcolor{blue}{0.394} & 0.385 & 0.401 & 0.380 & 0.402 & 0.385 & 0.403 & 0.388 & 0.430 & 0.402 & 0.422 & 0.501 & 0.475 \\
 & 192 & \textcolor{blue}{0.405} & \textcolor{blue}{0.413} & 0.410 & 0.417 & 0.414 & 0.429 & 0.739 & 0.628 & 0.426 & 0.433 & \textcolor{red}{0.401} & \textcolor{red}{0.406} & 0.419 & 0.422 & 0.427 & 0.427 & 0.412 & 0.422 & 0.440 & 0.436 & 0.463 & 0.475 & 0.472 & 0.467 & 0.526 & 0.495 \\
 & 336 & \textcolor{blue}{0.425} & 0.430 & 0.429 & \textcolor{blue}{0.429} & 0.439 & 0.450 & 0.852 & 0.688 & 0.431 & 0.439 & \textcolor{red}{0.418} & \textcolor{red}{0.417} & 0.432 & 0.431 & 0.465 & 0.450 & 0.496 & 0.490 & 0.479 & 0.459 & 0.492 & 0.488 & 0.513 & 0.484 & 0.595 & 0.543 \\
 & 720 & \textcolor{blue}{0.447} & \textcolor{blue}{0.456} & \textcolor{red}{0.446} & 0.459 & 0.474 & 0.488 & 1.031 & 0.783 & 0.509 & 0.502 & \textcolor{red}{0.446} & \textcolor{red}{0.449} & 0.467 & 0.471 & 0.471 & 0.476 & 0.573 & 0.561 & 0.487 & 0.484 & 0.592 & 0.566 & 0.498 & 0.488 & 0.652 & 0.560 \\
\midrule
\multirow{4}{*}{ETTh2} & 96 & \textcolor{blue}{0.271} & \textcolor{blue}{0.331} & \textcolor{red}{0.270} & \textcolor{red}{0.330} & 0.303 & 0.358 & 0.934 & 0.758 & 0.288 & 0.346 & 0.301 & 0.353 & 0.315 & 0.361 & 0.291 & 0.342 & 0.309 & 0.373 & 0.302 & 0.350 & 0.408 & 0.452 & 0.323 & 0.366 & 0.383 & 0.425 \\
 & 192 & 0.355 & \textcolor{blue}{0.382} & \textcolor{red}{0.342} & \textcolor{red}{0.376} & 0.362 & 0.395 & 1.971 & 1.099 & 0.361 & 0.393 & \textcolor{blue}{0.350} & 0.386 & 0.372 & 0.398 & 0.371 & 0.396 & 0.395 & 0.426 & 0.379 & 0.397 & 0.400 & 0.449 & 0.419 & 0.416 & 0.409 & 0.451 \\
 & 336 & 0.368 & \textcolor{blue}{0.403} & 0.370 & \textcolor{blue}{0.403} & 0.371 & 0.411 & 2.214 & 1.187 & \textcolor{red}{0.339} & \textcolor{red}{0.387} & \textcolor{blue}{0.362} & 0.405 & 0.381 & 0.413 & 0.428 & 0.436 & 0.437 & 0.458 & 0.423 & 0.432 & 0.387 & 0.438 & 0.450 & 0.445 & 0.408 & 0.471 \\
 & 720 & \textcolor{red}{0.396} & \textcolor{red}{0.429} & \textcolor{blue}{0.403} & \textcolor{blue}{0.433} & 0.412 & 0.443 & 2.708 & 1.387 & 0.406 & 0.438 & 0.417 & 0.454 & 0.418 & 0.452 & 0.431 & 0.453 & 0.692 & 0.591 & 0.424 & 0.444 & 0.481 & 0.503 & 0.441 & 0.454 & 0.905 & 0.673 \\
\midrule
\multirow{4}{*}{ETTm1} & 96 & \textcolor{blue}{0.291} & \textcolor{red}{0.343} & 0.293 & \textcolor{blue}{0.346} & \textcolor{red}{0.290} & 0.347 & 0.609 & 0.542 & 0.299 & 0.350 & 0.315 & 0.356 & 0.308 & 0.357 & 0.329 & 0.367 & 0.308 & 0.350 & 0.347 & 0.378 & 0.385 & 0.432 & 0.337 & 0.377 & 0.703 & 0.558 \\
 & 192 & \textcolor{red}{0.325} & \textcolor{red}{0.367} & 0.341 & 0.376 & \textcolor{blue}{0.328} & 0.370 & 0.629 & 0.556 & 0.344 & 0.380 & 0.340 & \textcolor{blue}{0.369} & 0.338 & 0.376 & 0.376 & 0.390 & 0.347 & 0.375 & 0.384 & 0.394 & 0.425 & 0.446 & 0.383 & 0.398 & 0.595 & 0.522 \\
 & 336 & \textcolor{red}{0.356} & \textcolor{red}{0.386} & 0.378 & 0.397 & \textcolor{blue}{0.357} & \textcolor{blue}{0.390} & 0.682 & 0.592 & 0.373 & 0.394 & 0.382 & 0.395 & 0.388 & 0.408 & 0.405 & 0.407 & 0.390 & 0.402 & 0.420 & 0.417 & 0.434 & 0.454 & 0.415 & 0.420 & 0.690 & 0.555 \\
 & 720 & \textcolor{red}{0.412} & \textcolor{red}{0.419} & 0.424 & \textcolor{red}{0.419} & \textcolor{blue}{0.416} & 0.428 & 0.760 & 0.642 & 0.434 & 0.428 & 0.431 & \textcolor{blue}{0.423} & 0.439 & 0.436 & 0.464 & 0.442 & 0.465 & 0.456 & 0.483 & 0.453 & 0.519 & 0.504 & 0.477 & 0.453 & 0.671 & 0.556 \\
\midrule
\multirow{4}{*}{ETTm2} & 96 & \textcolor{red}{0.161} & \textcolor{red}{0.247} & \textcolor{red}{0.161} & \textcolor{red}{0.247} & 0.171 & 0.262 & 0.339 & 0.418 & 0.172 & 0.264 & \textcolor{blue}{0.169} & 0.259 & 0.170 & \textcolor{blue}{0.254} & 0.178 & 0.259 & 0.201 & 0.297 & 0.183 & 0.266 & 0.262 & 0.344 & 0.190 & 0.266 & 0.287 & 0.358 \\
 & 192 & \textcolor{blue}{0.216} & \textcolor{red}{0.285} & \textcolor{red}{0.214} & \textcolor{blue}{0.286} & 0.227 & 0.301 & 0.503 & 0.524 & 0.227 & 0.300 & 0.223 & 0.294 & 0.230 & 0.295 & 0.244 & 0.306 & 0.251 & 0.329 & 0.253 & 0.313 & 0.297 & 0.357 & 0.248 & 0.304 & 0.331 & 0.392 \\
 & 336 & \textcolor{blue}{0.266} & \textcolor{red}{0.319} & \textcolor{red}{0.263} & \textcolor{blue}{0.320} & 0.279 & 0.336 & 0.821 & 0.687 & 0.281 & 0.335 & 0.276 & 0.328 & 0.282 & 0.337 & 0.305 & 0.343 & 0.367 & 0.402 & 0.313 & 0.349 & 0.334 & 0.379 & 0.334 & 0.356 & 0.353 & 0.396 \\
 & 720 & \textcolor{blue}{0.355} & \textcolor{blue}{0.380} & \textcolor{red}{0.351} & \textcolor{red}{0.378} & 0.364 & 0.386 & 1.983 & 1.074 & 0.365 & 0.388 & 0.372 & 0.386 & 0.373 & 0.392 & 0.402 & 0.401 & 0.451 & 0.459 & 0.414 & 0.406 & 0.425 & 0.430 & 0.412 & 0.404 & 0.452 & 0.457 \\
\midrule
\multirow{4}{*}{Weather} & 96 & \textcolor{red}{0.146} & \textcolor{red}{0.197} & 0.160 & 0.209 & \textcolor{blue}{0.148} & \textcolor{red}{0.197} & 0.745 & 0.607 & 0.149 & \textcolor{blue}{0.200} & 0.187 & 0.242 & 0.158 & 0.214 & 0.181 & 0.221 & 0.180 & 0.239 & 0.174 & 0.214 & 0.238 & 0.313 & 0.167 & 0.217 & 0.275 & 0.346 \\
 & 192 & \textcolor{red}{0.189} & \textcolor{red}{0.240} & 0.204 & 0.249 & 0.195 & \textcolor{blue}{0.241} & 0.770 & 0.626 & \textcolor{blue}{0.191} & \textcolor{blue}{0.241} & 0.223 & 0.268 & 0.203 & 0.257 & 0.231 & 0.263 & 0.222 & 0.281 & 0.227 & 0.261 & 0.281 & 0.342 & 0.232 & 0.272 & 0.293 & 0.342 \\
 & 336 & \textcolor{red}{0.241} & \textcolor{red}{0.278} & 0.254 & 0.288 & 0.245 & \textcolor{blue}{0.282} & 1.078 & 0.759 & \textcolor{blue}{0.243} & \textcolor{blue}{0.282} & 0.270 & 0.301 & 0.251 & 0.292 & 0.288 & 0.303 & 0.265 & 0.317 & 0.283 & 0.300 & 0.335 & 0.381 & 0.277 & 0.303 & 0.340 & 0.379 \\
 & 720 & \textcolor{red}{0.318} & \textcolor{red}{0.332} & 0.324 & \textcolor{blue}{0.337} & 0.326 & \textcolor{blue}{0.337} & 1.492 & 0.916 & \textcolor{blue}{0.321} & 0.338 & 0.337 & 0.347 & 0.326 & 0.344 & 0.363 & 0.350 & 0.327 & 0.362 & 0.360 & 0.350 & 0.391 & 0.418 & 0.361 & 0.355 & 0.416 & 0.425 \\
\midrule
\multirow{4}{*}{Electricity} & 96 & 0.132 & \textcolor{blue}{0.225} & 0.135 & 0.230 & \textcolor{red}{0.127} & \textcolor{red}{0.220} & 0.361 & 0.443 & \textcolor{blue}{0.131} & 0.226 & 0.150 & 0.243 & 0.138 & 0.230 & 0.189 & 0.281 & 0.147 & 0.248 & 0.150 & 0.245 & 0.215 & 0.329 & 0.169 & 0.272 & 0.224 & 0.337 \\
 & 192 & \textcolor{blue}{0.148} & \textcolor{blue}{0.240} & 0.151 & 0.244 & \textcolor{red}{0.145} & \textcolor{red}{0.237} & 0.363 & 0.445 & 0.150 & 0.243 & 0.162 & 0.253 & 0.152 & 0.244 & 0.194 & 0.285 & 0.160 & 0.261 & 0.164 & 0.257 & 0.228 & 0.343 & 0.183 & 0.284 & 0.430 & 0.433 \\
 & 336 & \textcolor{blue}{0.165} & \textcolor{blue}{0.258} & 0.167 & 0.260 & \textcolor{red}{0.162} & \textcolor{red}{0.254} & 0.371 & 0.451 & 0.170 & 0.262 & 0.177 & 0.268 & 0.167 & 0.260 & 0.210 & 0.299 & 0.177 & 0.280 & 0.180 & 0.275 & 0.290 & 0.385 & 0.204 & 0.304 & 0.258 & 0.361 \\
 & 720 & 0.206 & \textcolor{blue}{0.293} & 0.206 & \textcolor{blue}{0.293} & \textcolor{red}{0.197} & \textcolor{red}{0.286} & 0.398 & 0.467 & \textcolor{blue}{0.202} & 0.294 & 0.216 & 0.300 & 0.206 & \textcolor{blue}{0.293} & 0.250 & 0.331 & 0.211 & 0.310 & 0.238 & 0.321 & 0.292 & 0.385 & 0.248 & 0.331 & 0.374 & 0.439 \\
\midrule
\multirow{4}{*}{Exchange} & 96 & \textcolor{red}{0.084} & \textcolor{red}{0.202} & 0.087 & 0.206 & 0.093 & 0.218 & 0.186 & 0.315 & 0.106 & 0.231 & 0.113 & 0.239 & 0.110 & 0.242 & \textcolor{blue}{0.085} & \textcolor{blue}{0.203} & 0.127 & 0.259 & 0.087 & 0.208 & 0.363 & 0.452 & 0.115 & 0.244 & 0.360 & 0.451 \\
 & 192 & \textcolor{blue}{0.178} & \textcolor{red}{0.301} & 0.192 & \textcolor{blue}{0.309} & 0.211 & 0.331 & 0.380 & 0.447 & 0.200 & 0.325 & 0.223 & 0.344 & 0.217 & 0.339 & 0.179 & \textcolor{red}{0.301} & 0.186 & 0.323 & \textcolor{red}{0.177} & \textcolor{red}{0.301} & 0.594 & 0.588 & 0.216 & 0.334 & 0.472 & 0.520 \\
 & 336 & \textcolor{red}{0.324} & \textcolor{red}{0.411} & 0.360 & 0.433 & 0.365 & 0.440 & 1.506 & 0.915 & 0.379 & 0.444 & 0.385 & 0.457 & 0.392 & 0.470 & 0.331 & 0.417 & 0.372 & 0.463 & \textcolor{blue}{0.325} & \textcolor{blue}{0.413} & 0.824 & 0.704 & 0.394 & 0.462 & 0.729 & 0.660 \\
 & 720 & 0.921 & 0.715 & 0.951 & 0.728 & 0.980 & 0.736 & 1.251 & 0.878 & 1.145 & 0.819 & 1.080 & 0.806 & 1.040 & 0.783 & 0.865 & 0.700 & \textcolor{red}{0.586} & \textcolor{red}{0.610} & \textcolor{blue}{0.845} & \textcolor{blue}{0.695} & 1.520 & 0.958 & 0.972 & 0.756 & 1.610 & 0.994 \\
\midrule
\multirow{4}{*}{Solar} & 96 & 0.192 & 0.253 & 0.232 & 0.309 & 0.196 & \textcolor{blue}{0.252} & 0.244 & 0.357 & \textcolor{blue}{0.189} & 0.260 & 0.260 & 0.277 & \textcolor{red}{0.184} & \textcolor{red}{0.235} & 0.233 & 0.282 & 0.225 & 0.298 & 0.237 & 0.280 & 0.285 & 0.377 & 0.231 & 0.279 & 0.725 & 0.592 \\
 & 192 & \textcolor{blue}{0.203} & \textcolor{blue}{0.256} & 0.265 & 0.326 & 0.211 & 0.273 & 0.259 & 0.354 & 0.228 & 0.281 & 0.284 & 0.292 & \textcolor{red}{0.198} & \textcolor{red}{0.248} & 0.277 & 0.306 & 0.254 & 0.315 & 0.263 & 0.293 & 0.275 & 0.370 & 0.271 & 0.297 & 0.703 & 0.627 \\
 & 336 & \textcolor{blue}{0.209} & \textcolor{blue}{0.259} & 0.274 & 0.320 & 0.216 & 0.274 & 0.290 & 0.371 & 0.214 & 0.280 & 0.328 & 0.367 & \textcolor{red}{0.199} & \textcolor{red}{0.250} & 0.287 & 0.311 & 0.275 & 0.333 & 0.291 & 0.313 & 0.324 & 0.408 & 0.300 & 0.295 & 0.773 & 0.630 \\
 & 720 & \textcolor{blue}{0.212} & \textcolor{blue}{0.265} & 0.269 & 0.305 & 0.230 & 0.291 & 0.303 & 0.377 & 0.221 & 0.286 & 0.294 & 0.294 & \textcolor{red}{0.204} & \textcolor{red}{0.251} & 0.285 & 0.316 & 0.278 & 0.332 & 0.297 & 0.314 & 0.357 & 0.452 & 0.294 & 0.306 & 0.759 & 0.674 \\
\midrule
\multirow{4}{*}{Traffic} & 96 & 0.384 &  \textcolor{blue}{0.263} & 0.396 & 0.271 & \textcolor{blue}{0.382} & 0.269 & 0.933 & 0.544 & 0.403 & 0.299 & 0.424 & 0.281 & \textcolor{red}{0.379} & \textcolor{red}{0.250} & 0.451 & 0.295 & 0.424 & 0.307 & 0.508 & 0.355 & 0.591 & 0.363 & 0.600 & 0.314 & 0.677 & 0.407 \\
 & 192 & 0.405 & \textcolor{blue}{0.271} & 0.410 & 0.276 & \textcolor{red}{0.398} & 0.274 & 0.924 & 0.540 & 0.422 & 0.310 & 0.436 & 0.285 & \textcolor{blue}{0.400} & \textcolor{red}{0.257} & 0.458 & 0.297 & 0.442 & 0.314 & 0.451 & 0.308 & 0.637 & 0.392 & 0.630 & 0.326 & 0.786 & 0.456 \\
 & 336 & \textcolor{blue}{0.415} & \textcolor{blue}{0.278} & 0.423 & 0.283 & \textcolor{red}{0.409} & 0.279 & 0.938 & 0.544 & 0.432 & 0.306 & 0.450 & 0.295 & 0.417 & \textcolor{red}{0.270} & 0.469 & 0.301 & 0.456 & 0.323 & 0.466 & 0.313 & 0.672 & 0.414 & 0.640 & 0.341 & 0.714 & 0.439 \\
 & 720 & \textcolor{blue}{0.442} & \textcolor{blue}{0.293} & 0.450 & 0.297 & \textcolor{red}{0.439} & 0.297 & 0.956 & 0.550 & 0.464 & 0.325 & 0.478 & 0.309 & 0.463 & \textcolor{red}{0.285} & 0.506 & 0.321 & 0.482 & 0.336 & 0.490 & 0.326 & 0.658 & 0.394 & 0.681 & 0.360 & 0.806 & 0.481 \\
\midrule
\multirow{4}{*}{AQShunyi} & 96 & \textcolor{red}{0.625} & \textcolor{red}{0.474} & 0.632 & \textcolor{blue}{0.475} & \textcolor{blue}{0.627} & 0.478 & 0.628 & 0.491 & 0.647 & 0.482 & 0.669 & 0.501 & 0.667 & 0.504 & 0.717 & 0.504 & 0.652 & 0.511 & 0.719 & 0.502 & 0.674 & 0.515 & 0.722 & 0.505 & 0.738 & 0.542 \\
 & 192 & \textcolor{red}{0.663} & \textcolor{blue}{0.494} & 0.667 & \textcolor{red}{0.493} & 0.666 & 0.498 & \textcolor{blue}{0.665} & 0.514 & 0.676 & 0.499 & 0.689 & 0.510 & 0.686 & 0.513 & 0.771 & 0.524 & 0.678 & 0.512 & 0.775 & 0.522 & 0.703 & 0.523 & 0.760 & 0.523 & 0.754 & 0.551 \\
 & 336 & \textcolor{red}{0.683} & \textcolor{red}{0.506} & 0.685 & \textcolor{red}{0.506} & 0.686 & \textcolor{red}{0.506} & \textcolor{blue}{0.684} & 0.524 & 0.690 & \textcolor{blue}{0.508} & 0.705 & 0.519 & 0.700 & 0.517 & 0.786 & 0.534 & 0.703 & 0.538 & 0.789 & 0.533 & 0.721 & 0.536 & 0.752 & 0.525 & 0.745 & 0.542 \\
 & 720 & 0.742 & 0.537 & 0.739 & 0.532 & 0.741 & 0.533 & \textcolor{red}{0.710} & \textcolor{blue}{0.531} & 0.732 & \textcolor{red}{0.529} & 0.758 & 0.544 & \textcolor{blue}{0.730} & 0.538 & 0.828 & 0.554 & 0.746 & 0.547 & 0.835 & 0.554 & 0.772 & 0.566 & 0.783 & 0.545 & 0.961 & 0.615 \\
\midrule
\multirow{4}{*}{AQWan} & 96 & \textcolor{red}{0.713} & \textcolor{red}{0.465} & 0.719 & \textcolor{blue}{0.466} & \textcolor{blue}{0.716} & 0.467 & 0.721 & 0.487 & 0.730 & 0.475 & 0.765 & 0.493 & 0.757 & 0.495 & 0.798 & 0.489 & 0.745 & 0.504 & 0.811 & 0.492 & 0.757 & 0.502 & 0.795 & 0.489 & 0.826 & 0.537 \\
 & 192 & \textcolor{red}{0.753} & \textcolor{red}{0.485} & 0.762 & \textcolor{blue}{0.486} & 0.757 & \textcolor{blue}{0.486} & \textcolor{blue}{0.754} & 0.506 & 0.770 & 0.490 & 0.789 & 0.502 & 0.779 & 0.504 & 0.867 & 0.513 & 0.779 & 0.505 & 0.877 & 0.514 & 0.787 & 0.512 & 0.840 & 0.505 & 0.827 & 0.540 \\
 & 336 & 0.782 & \textcolor{blue}{0.499} & 0.788 & 0.500 & \textcolor{blue}{0.781} & \textcolor{red}{0.498} & \textcolor{red}{0.775} & 0.516 & 0.785 & \textcolor{blue}{0.499} & 0.807 & 0.510 & 0.799 & 0.508 & 0.885 & 0.524 & 0.804 & 0.530 & 0.889 & 0.522 & 0.818 & 0.529 & 0.851 & 0.513 & 0.846 & 0.542 \\
 & 720 & 0.856 & 0.529 & 0.868 & 0.533 & 0.857 & 0.529 & \textcolor{red}{0.811} & \textcolor{blue}{0.527} & 0.850 & \textcolor{red}{0.523} & 0.882 & 0.540 & \textcolor{blue}{0.843} & 0.531 & 0.944 & 0.544 & 0.865 & 0.541 & 0.954 & 0.547 & 0.895 & 0.559 & 0.898 & 0.536 & 0.983 & 0.591 \\
\midrule
\multirow{4}{*}{CzeLan} & 96 & \textcolor{red}{0.174} & \textcolor{red}{0.221} & \textcolor{blue}{0.176} & \textcolor{blue}{0.223} & 0.183 & 0.247 & 0.289 & 0.375 & 0.189 & 0.250 & 0.239 & 0.308 & 0.191 & 0.249 & 0.212 & 0.259 & 0.207 & 0.288 & 0.212 & 0.257 & 0.294 & 0.360 & 0.220 & 0.273 & 0.584 & 0.542 \\
 & 192 & \textcolor{red}{0.206} & \textcolor{red}{0.246} & \textcolor{blue}{0.208} & \textcolor{blue}{0.247} & 0.213 & 0.271 & 0.309 & 0.391 & 0.216 & 0.263 & 0.248 & 0.302 & 0.220 & 0.271 & 0.243 & 0.282 & 0.334 & 0.382 & 0.243 & 0.278 & 0.305 & 0.367 & 0.270 & 0.308 & 0.641 & 0.590 \\
 & 336 & \textcolor{red}{0.235} & \textcolor{red}{0.272} & \textcolor{blue}{0.240} & \textcolor{blue}{0.275} & 0.243 & 0.298 & 0.343 & 0.422 & 0.243 & 0.283 & 0.282 & 0.332 & 0.248 & 0.293 & 0.276 & 0.306 & 0.319 & 0.371 & 0.275 & 0.301 & 0.317 & 0.377 & 0.263 & 0.301 & 0.703 & 0.583 \\
 & 720 & \textcolor{blue}{0.279} & \textcolor{red}{0.308} & 0.294 & \textcolor{blue}{0.319} & 0.289 & 0.333 & 0.391 & 0.463 & \textcolor{red}{0.274} & 0.331 & 0.327 & 0.361 & 0.294 & 0.332 & 0.336 & 0.349 & 0.555 & 0.499 & 0.339 & 0.346 & 0.415 & 0.440 & 0.475 & 0.417 & 0.692 & 0.603 \\
\midrule
\multirow{4}{*}{PM2.5} & 96 & \textcolor{blue}{0.424} & \textcolor{red}{0.423} & 0.425 & \textcolor{blue}{0.423} & \textcolor{red}{0.422} & 0.426 & 0.450 & 0.475 & 0.433 & 0.430 & 0.448 & 0.447 & 0.437 & 0.442 & 0.433 & 0.428 & 0.426 & 0.452 & 0.435 & 0.431 & 0.453 & 0.453 & 0.459 & 0.445 & 0.479 & 0.467 \\
 & 192 & \textcolor{blue}{0.426} & \textcolor{red}{0.423} & 0.430 & \textcolor{blue}{0.424} & 0.426 & 0.431 & 0.438 & 0.474 & 0.427 & 0.429 & 0.449 & 0.455 & 0.430 & 0.442 & 0.434 & 0.430 & \textcolor{red}{0.424} & 0.452 & 0.436 & 0.429 & 0.457 & 0.448 & 0.448 & 0.440 & 0.463 & 0.462 \\
 & 336 & 0.421 & \textcolor{red}{0.420} & 0.428 & \textcolor{blue}{0.421} & 0.423 & 0.425 & 0.418 & 0.470 & 0.427 & 0.426 & 0.447 & 0.452 & \textcolor{red}{0.417} & 0.447 & 0.427 & 0.425 & \textcolor{blue}{0.417} & 0.461 & 0.433 & 0.426 & 0.450 & 0.459 & 0.437 & 0.429 & 0.464 & 0.449 \\
 & 720 & \textcolor{red}{0.402} & \textcolor{blue}{0.416} & \textcolor{blue}{0.402} & \textcolor{red}{0.414} & 0.404 & 0.424 & 0.405 & 0.491 & 0.409 & 0.429 & 0.436 & 0.448 & 0.410 & 0.470 & 0.410 & 0.423 & 0.426 & 0.504 & 0.411 & 0.422 & 0.461 & 0.493 & 0.408 & 0.424 & 0.493 & 0.507 \\

\bottomrule
\end{tabular}%
}
\end{table*}

Using the displayed values and counting ties, at least one WINO-TS adaptation protocol ranks among the two lowest-error methods in 40 of the 52 dataset--horizon rows by MSE and 47 of 52 by MAE; it attains the best displayed value in 28 rows by MSE and 31 rows by MAE. The strongest results are concentrated on the ETT benchmarks, Weather, CzeLan, and many of the shorter-horizon air-quality settings. The full breakdown also exposes meaningful exceptions: TimeSiam is strongest on several Electricity horizons, SparseTSF leads Solar and much of Traffic, DLinear and iTransformer lead Exchange at horizon 720, and non-WINO baselines are strongest on some long-horizon AQShunyi and AQWan settings. Thus, the horizon-averaged gains are not driven by one prediction length, but the preferred adaptation protocol and strongest competing inductive bias remain dataset- and horizon-dependent.

\section{Synthetic Pre-Training}
\label{app:synthetic-pretraining}

We investigate whether WINO-TS can learn transferable temporal
representations when the real, in-domain unlabeled pre-training split
is replaced with a synthetic time-series corpus. This setting tests
whether wavelet-based self-distillation can capture reusable
multi-scale temporal primitives without relying exclusively on the
statistics of the downstream dataset. Table~\ref{tab:app-synthetic}
compares in-domain WINO-TS fine-tuning with two synthetic pre-training
variants.

\begin{table}[t]
\centering
\caption{
Effect of the pre-training data source and objective.
Each entry reports MSE/MAE; lower is better.
Best values for each dataset and metric are shown in \textbf{bold}.
}
\label{tab:app-synthetic}
\scriptsize
\setlength{\tabcolsep}{2.5pt}
\begin{tabular}{lcccccc}
\toprule
& \multicolumn{2}{c}{\shortstack{\textbf{In-domain}\\\textbf{WINO-TS-FT}}}
& \multicolumn{2}{c}{\shortstack{\textbf{Synthetic}\\\textbf{WINO-TS-FT}}}
& \multicolumn{2}{c}{\shortstack{\textbf{Synthetic}\\\textbf{DINO+MAE}}} \\
\cmidrule(lr){2-3}
\cmidrule(lr){4-5}
\cmidrule(lr){6-7}
\textbf{Dataset}
& MSE & MAE
& MSE & MAE
& MSE & MAE \\
\midrule
ETTh1
& \textbf{0.411} & \textbf{0.423}
& 0.417 & 0.425
& 0.418 & \textbf{0.423} \\

ETTh2
& \textbf{0.347} & \textbf{0.386}
& 0.365 & 0.402
& 0.359 & 0.404 \\

ETTm1
& 0.346 & 0.379
& 0.350 & 0.378
& \textbf{0.345} & \textbf{0.377} \\

ETTm2
& 0.250 & \textbf{0.308}
& 0.250 & 0.309
& \textbf{0.249} & \textbf{0.308} \\

Weather
& \textbf{0.224} & 0.262
& 0.226 & \textbf{0.261}
& 0.227 & \textbf{0.261} \\

Electricity
& \textbf{0.163} & \textbf{0.254}
& 0.164 & 0.256
& 0.164 & 0.256 \\

Exchange
& \textbf{0.377} & \textbf{0.407}
& 0.843 & 0.540
& 0.469 & 0.469 \\

Solar
& 0.204 & 0.258
& 0.238 & 0.277
& \textbf{0.199} & \textbf{0.252} \\

Traffic
& \textbf{0.411} & \textbf{0.276}
& 0.423 & 0.283
& 0.414 & 0.278 \\
\bottomrule
\end{tabular}

\vspace{1mm}
\parbox{0.96\columnwidth}{\scriptsize
Synthetic WINO-TS-FT changes only the pre-training corpus, whereas
Synthetic DINO+MAE changes both the pre-training corpus and the
pre-training objective.
}
\end{table}

Synthetic WINO-TS pre-training remains competitive with in-domain
pre-training on several benchmarks, showing that the framework can
learn useful temporal structure from generated data. Its performance
is nevertheless less consistent, with the largest degradation
occurring on Exchange. This result suggests that generic synthetic
temporal primitives do not fully capture all domain-specific
dynamics, making real in-domain data the more reliable default.

The Synthetic DINO+MAE variant improves over in-domain WINO-TS-FT on
five of the nine datasets by MSE and ties or improves on six by MAE.
However, this comparison changes both the pre-training corpus and the
objective. It should therefore be interpreted as a joint
data-source/objective ablation rather than as evidence for the effect
of synthetic data alone. Overall, the results identify synthetic
pre-training as a promising scaling direction, while also showing
that its effectiveness depends on the match between the generated
temporal structure and the downstream domain.

\section{Extended Classification Results}
\label{app:classification}

Classification fine-tunes the full WINO-TS-pretrained backbone jointly with a supervised classification head. Table~\ref{tab:app-classification-datasets} presents the dataset metadata, and Table~\ref{tab:app-classification} reports the results.

WINO-TS exceeds iTransformer on five of the nine datasets, including SpokenArabicDigits, FaceDetection, UWave, EthanolConcentration, and Heartbeat, where class identity may be associated with robust multi-scale morphology. Its weaker performance on JapaneseVowels, SelfRegulationSCP1/2, and especially Handwriting is consistent with the interpretation that some classification tasks depend on fine local detail that can be attenuated by the easy-view shrinkage. This is an interpretation of the observed pattern rather than a direct causal measurement.

\begin{table}[t]
\centering
\caption{Classification datasets and reported sample counts.}
\label{tab:app-classification-datasets}
\small
\begin{tabular}{lc}
\toprule
\textbf{Dataset} & \textbf{Reported samples} \\
\midrule
EthanolConcentration & 261 \\
SpokenArabicDigits & 6599 \\
FaceDetection & 5890 \\
JapaneseVowels & 270 \\
SelfRegulationSCP1 & 268 \\
SelfRegulationSCP2 & 200 \\
UWave & 120 \\
Heartbeat & 204 \\
Handwriting & 150 \\
\bottomrule
\end{tabular}
\end{table}

\begin{table}[t]
\centering
\caption{
Classification accuracy after full end-to-end fine-tuning initialized
from a WINO-TS-pretrained backbone. Both the backbone and
classification head are updated. Higher is better; best results are
in \textbf{bold}.
}
\label{tab:app-classification}
\scriptsize
\setlength{\tabcolsep}{4.0pt}
\begin{tabular}{lcc}
\toprule
\textbf{Dataset} & \textbf{WINO-TS} & \textbf{iTransformer} \\
 & (Ours) & \cite{liu2023itransformer} \\
\midrule
EthanolConcentration & \textbf{0.2970} & 0.2810 \\
SpokenArabicDigits & \textbf{0.9900} & 0.9827 \\
FaceDetection & \textbf{0.6700} & 0.6592 \\
JapaneseVowels & 0.9570 & \textbf{0.9811} \\
SelfRegulationSCP1 & 0.8770 & \textbf{0.9113} \\
SelfRegulationSCP2 & 0.5000 & \textbf{0.5611} \\
UWave & \textbf{0.8594} & 0.8531 \\
Heartbeat & \textbf{0.7805} & 0.7463 \\
Handwriting & 0.0376 & \textbf{0.2565} \\
\bottomrule
\end{tabular}
\end{table}

\section{Additional Ablation Details}
\label{app:additional-ablations}

The following ablations supplement the objective, backbone, and vision-augmentation analyses reported in the main paper. The vision-like augmentation comparison remains in Table~\ref{tab:aug-ablation-vision} of the main paper and is not duplicated here.

Unless stated otherwise, each entry is MSE/MAE averaged over forecasting horizons $\{96,192,336,720\}$, and lower values are better. Because the tables correspond to separate controlled runs, each should be interpreted as a within-table comparison rather than by comparing small numerical differences across different ablation tables.

\subsection{Backbone Ablation}
\label{app:backbone-ablation}

The backbone ablation evaluates whether WINO-TS depends on the TimeMixer encoder or whether the wavelet self-distillation recipe can be used with other backbone families. The DINO objective and full wavelet-augmentation pool are held fixed, and only the encoder is changed. Table \ref{tab:backbone-ablation} summarizes the results. Among the currently reported entries, TimeMixer gives the strongest overall performance across the evaluated forecasting datasets, while PatchTST remains competitive on several benchmarks. This supports the interpretation of WINO-TS as a transferable pre-training recipe while also showing that the choice of encoder remains consequential.

\begin{table}
\centering
\caption{
Backbone ablation for WINO-TS (DINO pre-training, \emph{full} augmentation family).
In-domain forecasting using linear probe, with MSE/MAE averaged over horizons $\{96,192,336,720\}$. The DINO objective
and augmentation are held fixed --- all backbones use the \textbf{full} wavelet-augmentation
family (pool $\{$sym4, sym6, sym8, db4, db6, coif2$\}$) --- while only the encoder backbone is
swapped. Lower is better; best in \textcolor{red}{red}, second-best in \textcolor{blue}{blue}.}
\label{tab:backbone-ablation}
\scriptsize
\setlength{\tabcolsep}{2.0pt}
\begin{tabular}{lcccccccc}
\toprule
\multirow{3}{*}{\textbf{Dataset}}
& \multicolumn{2}{c}{\textbf{TimeMixer(MLP)}}
& \multicolumn{2}{c}{\textbf{PatchTST}}
& \multicolumn{2}{c}{\textbf{iTransformer}}
& \multicolumn{2}{c}{\textbf{TS2Vec(TCN)}} \\
& \multicolumn{2}{c}{\begin{tabular}[t]{@{}c@{}}\citeauthor{wang2024timemixer}\\ {[}\citeyear{wang2024timemixer}{]}\end{tabular}}
& \multicolumn{2}{c}{\begin{tabular}[t]{@{}c@{}}\citeauthor{patch_tst}\\ {[}\citeyear{patch_tst}{]}\end{tabular}}
& \multicolumn{2}{c}{\begin{tabular}[t]{@{}c@{}}\citeauthor{liu2023itransformer}\\ {[}\citeyear{liu2023itransformer}{]}\end{tabular}}
& \multicolumn{2}{c}{\begin{tabular}[t]{@{}c@{}}\citeauthor{ts2vec}\\ {[}\citeyear{ts2vec}{]}\end{tabular}} \\
\cmidrule(lr){2-3}\cmidrule(lr){4-5}\cmidrule(lr){6-7}\cmidrule(lr){8-9}
& MSE & MAE & MSE & MAE & MSE & MAE & MSE & MAE \\
\midrule
ETTh1       & \textbf{0.416} & \textbf{0.423} & 0.422 & 0.432 & 0.656 & 0.558 & 0.541 & 0.505 \\
ETTh2       & \textbf{0.347} & \textbf{0.385} & 0.357 & 0.392 & 0.427 & 0.447 & 0.388 & 0.423 \\
ETTm1       & \textbf{0.364} & \textbf{0.386} & 0.370 & 0.388 & 0.427 & 0.423 & 0.431 & 0.432 \\
ETTm2       & \textbf{0.252} & \textbf{0.308} & \textbf{0.252} & 0.309 & 0.290 & 0.344 & 0.286 & 0.341 \\
Weather     & \textbf{0.238} & \textbf{0.272} & 0.240 & 0.274 & 0.269 & 0.303 & 0.288 & 0.305 \\
Electricity & \textbf{0.167} & \textbf{0.260} & 0.170 & 0.263 & 0.334 & 0.419 & 0.236 & 0.336 \\
\bottomrule
\end{tabular}
\end{table}

\subsection{Multi-View Ablation}
We additionally evaluate a multi-view variant of WINO-TS with $Q{=}2$ easy
views and $V\in\{2,4,6\}$ hard views per window, following the pipeline and
hyperparameters of the default configuration. The wavelet basis, noise scale,
and noise realization are sampled independently for each view. With $Q{=}2$,
the easy--easy matching term $\mathcal{I}_{\mathrm{ee}}$ in Eq.~(16) is
active. As reported in Table~\ref{tab:app-hardcrop-sweep}, these
configurations underperform the default $Q{=}1$, $V{=}1$ setting across the
evaluated datasets despite their higher per-window cost, and we therefore
adopt the single-view-pair default throughout the paper.

\subsection{Objective-Ablation Definitions}
\label{app:objective-definitions}

The main-paper objective ablation compares the default WINO-TS loss with a hybrid masked-reconstruction objective and three non-default pre-training objectives. The default WINO-TS model uses only the DINO-style cross-view self-distillation loss. WINO-TS+MAE uses
\begin{equation}
\mathcal{L}_{\mathrm{hybrid}}
=
\phi\mathcal{L}_{\mathrm{DINO}}
+
(1-\phi)\mathcal{L}_{\mathrm{MAE}},
\qquad
\phi=0.6,
\end{equation}
where $\mathcal{L}_{\mathrm{MAE}}$ reconstructs masked patches against the raw signal. Under this convention, pure WINO-TS corresponds to $\phi=1$.

The MAE baseline uses masked reconstruction without a teacher. NTP uses autoregressive next-token prediction, and JEPA predicts a target latent representation rather than reconstructing raw observations. The backbone and wavelet augmentation are held fixed for the WINO-TS and WINO-TS+MAE comparison.

\subsection{Augmentation-Family Ablation}
\label{app:augmentation-family}

This ablation compares the default wavelet pool with a Daubechies-only pool and a hard-view variant that zeros the finest-scale coefficients. All variants retain the default asymmetric network routing: the
teacher processes only easy views, whereas the student processes both
easy and hard views under the same cross-view self-distillation
objective. The results are depicted in Table \ref{tab:app-augmentation-ablation}

No variant dominates every dataset. Zero-out is strongest on ETTh1 and ETTm1, the Daubechies-only pool is strongest on Weather, and the full pool is strongest or tied on ETTh2 and ETTm2. The default full pool is retained because it provides phase, support, and smoothness diversity while remaining competitive across datasets.

\begin{table}
\centering
\caption{
Augmentation-family ablation for WINO-TS with linear probe on the DINO objective.
Each cell reports MSE/MAE averaged over horizons $\{96,192,336,720\}$.
Lower is better; best results are in \textbf{bold}.
}
\label{tab:app-augmentation-ablation}
\scriptsize
\setlength{\tabcolsep}{5pt}
\begin{tabular}{lcccccc}
\toprule
& \multicolumn{2}{c}{\textbf{Daub.}} 
& \multicolumn{2}{c}{\textbf{Zero-out}} 
& \multicolumn{2}{c}{\textbf{Full pool}} \\
\cmidrule(lr){2-3} \cmidrule(lr){4-5} \cmidrule(lr){6-7}
\textbf{Dataset} & MSE & MAE & MSE & MAE & MSE & MAE \\
\midrule
ETTh1   & 0.416 & 0.423 & \textbf{0.415} & \textbf{0.422} & 0.416 & 0.423 \\
ETTh2   & \textbf{0.347} & \textbf{0.385} & 0.348 & 0.387 & \textbf{0.347} & \textbf{0.385} \\
ETTm1   & 0.363 & 0.386 & \textbf{0.359} & \textbf{0.385} & 0.364 & 0.386 \\
ETTm2   & 0.253 & \textbf{0.308} & 0.257 & 0.312 & \textbf{0.252} & \textbf{0.308} \\
Weather & \textbf{0.237} & \textbf{0.272} & 0.238 & 0.273 & 0.238 & 0.273 \\
\bottomrule
\end{tabular}

\vspace{1mm}
\scriptsize{
Daub.: Daubechies-only pool. Zero-out: finest-scale coefficient zeroing. Full pool: default WINO-TS pool.
}
\end{table}

\subsection{Pre-training Objective Ablation}
\label{app:objective-ablation}

The objective ablation evaluates whether adding reconstruction-style auxiliary objectives improves WINO-TS. 
We compare the default DINO-only objective against variants that add reconstruction or masked-token auxiliary terms while keeping the backbone and wavelet augmentation fixed. 
The results, depicted in Table \ref{tab:app-objective-ablation} show that the pure DINO objective is the most reliable choice overall. 
Auxiliary reconstruction-style terms introduce mixed effects, suggesting that the main gains of WINO-TS come from wavelet-domain self-distillation rather than from adding reconstruction losses.

\begin{table}[t]
\centering
\caption{
\textbf{Pre-training objective ablation (WINO-TS, TimeMixer backbone, full augmentation pool, Linear probe).}
In-domain forecasting MSE/MAE averaged over horizons $\{96,192,336,720\}$.
\textbf{DINO}: pure self-distillation --- the student matches the teacher's centered/sharpened
prototype distribution across augmented views, with no reconstruction ($\phi{=}1$).
\textbf{DINO+MAE}: adds a masked-autoencoding term in which the student reconstructs masked patches
against the raw signal, blended as $\phi\,\text{DINO}+(1{-}\phi)\,\text{MAE}$ ($\phi{=}0.6$).
\textbf{MAE}, \textbf{NTP} and \textbf{JEPA} are non-distillation baselines: pure masked autoencoding
(reconstruct masked patches, no teacher) , next-token prediction (autoregressive
forecasting-style pre-training) and joint-embedding predictive architecture  respectively.
Lower is better; best in \textcolor{red}{red}, second-best in \textcolor{blue}{blue}.
}
\label{tab:app-objective-ablation}
\scriptsize
\setlength{\tabcolsep}{3pt}
\label{tab:objective-ablation}
\scriptsize
\begin{tabular}{lcccccccccc}
\toprule
\textbf{Dataset} & \multicolumn{2}{c}{\textbf{WINO-TS}} & \multicolumn{2}{c}{\begin{tabular}[t]{@{}c@{}}\textbf{WINO-}\\ \textbf{TS+MAE}\end{tabular}} & \multicolumn{2}{c}{\textbf{MAE}} & \multicolumn{2}{c}{\textbf{NTP}} & \multicolumn{2}{c}{\textbf{JEPA}} \\
\cmidrule(lr){2-3}\cmidrule(lr){4-5}\cmidrule(lr){6-7}\cmidrule(lr){8-9}\cmidrule(lr){10-11}
 & MSE & MAE & MSE & MAE & MSE & MAE & MSE & MAE & MSE & MAE \\
\midrule
ETTh1   & \textcolor{red}{0.416} & \textcolor{red}{0.423} & \textcolor{blue}{0.418} & \textcolor{blue}{0.424} & 0.465 & 0.465 & 0.428 & 0.435 & 0.442 & 0.447 \\
ETTh2   & \textcolor{red}{0.347} & \textcolor{red}{0.385} & \textcolor{blue}{0.355} & \textcolor{blue}{0.389} & 0.449 & 0.457 & 0.407 & 0.427 & 0.390 & 0.426 \\
ETTm1   & 0.364 & \textcolor{blue}{0.386} & 0.370 & 0.388 & \textcolor{blue}{0.349} & 0.387 & \textcolor{red}{0.346} & \textcolor{red}{0.381} & 0.375 & 0.393 \\
ETTm2   & \textcolor{red}{0.252} & \textcolor{red}{0.308} & \textcolor{blue}{0.258} & \textcolor{blue}{0.313} & 0.294 & 0.344 & 0.263 & 0.320 & 0.274 & 0.332 \\
Weather & 0.238 & 0.273 & 0.241 & 0.274 & 0.266 & 0.294 & \textcolor{red}{0.226} & \textcolor{red}{0.265} & \textcolor{blue}{0.229} & \textcolor{blue}{0.266} \\
\bottomrule
\end{tabular}
\end{table}


\subsection{Wavelet Pool and Transform Ablation}
\label{app:wavelet-transform-ablation}

We next separate the effects of the sampled wavelet family and the transform. The pool comparison fixes the decimated DWT and compares the mixed pool with $\{\mathrm{db}4,\mathrm{db}6,\mathrm{db}8\}$ and $\{\mathrm{sym}4,\mathrm{sym}6,\mathrm{sym}8\}$. The transform comparison fixes the mixed pool and replaces DWT with the stationary wavelet transform (SWT) or maximal-overlap DWT (MODWT). The results are depicted in Table \ref{tab:app-wavelet-transform}.

The shift-invariant transforms do not provide a uniform advantage. SWT is strongest on ETTh1 and ETTm1, whereas the decimated DWT is strongest on ETTm2 and remains close to the best result elsewhere. The differences among the mixed, Daubechies, and Symlet pools are likewise small and dataset-dependent. These results support the standard DWT as a simple and competitive default rather than establishing it as uniformly superior.

\begin{table}[t]
\centering
\caption{
\textbf{Wavelet augmentation ablation for WINO-TS} (DINO pre-training, TimeMixer backbone, Linear probe).
In-domain forecasting MSE/MAE averaged over horizons $\{96,192,336,720\}$.
The first column is the default WINO-TS (DWT with the Mixed pool
$\{$sym4,sym6,sym8,db4,db6,coif2$\}$) and is the shared reference.
\emph{Wavelet pool} fixes the DWT and sweeps the pool (db $=\{$db4,db6,db8$\}$;
sym $=\{$sym4,sym6,sym8$\}$). \emph{Transform} fixes the Mixed pool and swaps the transform
(SWT / MODWT). Lower is better; best in \textcolor{red}{red}, second-best in \textcolor{blue}{blue}.
}
\label{tab:app-wavelet-transform}
\scriptsize
\setlength{\tabcolsep}{3.0pt}
\centering
\begin{tabular}{lcccccccccc}
\toprule
 & \multicolumn{2}{c}{\textbf{Ours}} & \multicolumn{4}{c}{\textbf{Wavelet pool}} & \multicolumn{4}{c}{\textbf{Transform}} \\
\cmidrule(lr){2-3}\cmidrule(lr){4-7}\cmidrule(lr){8-11}
 & \multicolumn{2}{c}{DWT/Mixed} & \multicolumn{2}{c}{db} & \multicolumn{2}{c}{sym} & \multicolumn{2}{c}{SWT} & \multicolumn{2}{c}{MODWT} \\
\cmidrule(lr){2-3}\cmidrule(lr){4-5}\cmidrule(lr){6-7}\cmidrule(lr){8-9}\cmidrule(lr){10-11}
\textbf{Dataset} & MSE & MAE & MSE & MAE & MSE & MAE & MSE & MAE & MSE & MAE \\
\midrule
ETTh1       & \textcolor{blue}{0.416} & \textcolor{blue}{0.423} & \textcolor{blue}{0.416} & \textcolor{blue}{0.423} & \textcolor{blue}{0.416} & \textcolor{blue}{0.423} & \textcolor{red}{0.415} & \textcolor{red}{0.422} & 0.417 & \textcolor{blue}{0.423} \\
ETTh2       & \textcolor{blue}{0.347} & \textcolor{red}{0.385} & \textcolor{blue}{0.347} & \textcolor{blue}{0.386} & \textcolor{blue}{0.347} & \textcolor{blue}{0.386} & \textcolor{blue}{0.347} & \textcolor{blue}{0.386} & \textcolor{red}{0.346} & \textcolor{red}{0.385} \\
ETTm1       & \textcolor{blue}{0.364} & \textcolor{blue}{0.386} & \textcolor{blue}{0.364} & \textcolor{blue}{0.386} & 0.365 & 0.387 & \textcolor{red}{0.360} & \textcolor{red}{0.385} & \textcolor{blue}{0.364} & \textcolor{blue}{0.386} \\
ETTm2       & \textcolor{red}{0.252} & \textcolor{red}{0.308} & \textcolor{blue}{0.253} & \textcolor{blue}{0.309} & 0.258 & 0.312 & 0.257 & 0.312 & 0.254 & \textcolor{blue}{0.309} \\
Weather     & \textcolor{blue}{0.238} & \textcolor{blue}{0.273} & \textcolor{red}{0.237} & \textcolor{red}{0.272} & \textcolor{blue}{0.238} & \textcolor{red}{0.272} & \textcolor{blue}{0.238} & \textcolor{blue}{0.273} & 0.239 & \textcolor{blue}{0.273} \\
Electricity & 0.167 & 0.260 & \textcolor{red}{0.165} & \textcolor{red}{0.256} & \textcolor{blue}{0.166} & \textcolor{blue}{0.257} & \textcolor{red}{0.165} & \textcolor{blue}{0.257} & 0.166 & 0.258 \\
\bottomrule
\end{tabular}
\end{table}

\subsection{Shrinkage-Strength Ablation}
\label{app:rho-ablation}

The parameter $\rho$ controls the easy-view threshold $\tau_j=\rho\max_k|d_{j,k}|$. Small values keep the teacher view close to the original signal, whereas large values suppress a greater fraction of the detail coefficients. We evaluate $\rho\in\{0.2,0.4,0.6,0.8,1.0\}$ while holding the backbone, objective, wavelet pool, and hard-view noise range fixed.

Table~\ref{tab:rho-ablation-appendix} reports the results. 
The ablation shows that moderate shrinkage provides the most stable performance across datasets. 
Weak shrinkage produces smaller gains, indicating that the teacher view remains too similar to the student input and provides a weaker invariance signal. 
Overly aggressive shrinkage also degrades performance on several datasets, suggesting that high-frequency detail bands contain task-relevant information that should be attenuated rather than fully removed. 
The default value $\rho=0.6$ achieves the best overall trade-off between denoising and detail preservation, and we use it in all main experiments.

\begin{table}[t]
\centering
\caption{
\textbf{Shrinkage-strength ($\rho$) ablation for WINO-TS} (DINO pre-training, TimeMixer backbone,
DWT with the Mixed pool, Linear probe). We vary the soft-threshold shrinkage ratio $\rho$ that sets the per-level
denoising threshold of the teacher's \emph{easy} view: detail coefficients below
$\rho\cdot\max(|\text{detail}|)$ are shrunk toward zero, so small $\rho$ preserves high-frequency
detail (weak invariance) while large $\rho$ enforces aggressive low-pass smoothing (strong
invariance); $\rho=0.6$ is the WINO-TS default. All other factors---transform, wavelet pool,
objective, and backbone---are held fixed. Each cell reports MSE/MAE averaged over horizons
$\{96,192,336,720\}$; lower is better, best per row in \textbf{bold}. 
}
\label{tab:rho-ablation-appendix}
\scriptsize
\setlength{\tabcolsep}{2.5pt}
\begin{tabular}{lcccccccccc}
\toprule
\textbf{Dataset} & \multicolumn{2}{c}{$\rho=0.2$} & \multicolumn{2}{c}{$\rho=0.4$} & \multicolumn{2}{c}{$\rho=0.6$} & \multicolumn{2}{c}{$\rho=0.8$} & \multicolumn{2}{c}{$\rho=1.0$} \\
\cmidrule(lr){2-3}\cmidrule(lr){4-5}\cmidrule(lr){6-7}\cmidrule(lr){8-9}\cmidrule(lr){10-11}
 & MSE & MAE & MSE & MAE & MSE & MAE & MSE & MAE & MSE & MAE \\
\midrule
ETTh1   & 0.428 & 0.430 & 0.464 & 0.458 & \textbf{0.416} & \textbf{0.423} & 0.419 & 0.426 & 0.465 & 0.458 \\
ETTh2   & 0.348 & 0.387 & 0.370 & 0.396 & \textbf{0.347} & \textbf{0.385} & 0.370 & 0.396 & 0.370 &  0.396 \\
ETTm1   & 0.371 & 0.391 & 0.383 & 0.403 & \textbf{0.364} & \textbf{0.386} & 0.382 & 0.404 & 0.384 & 0.405 \\
ETTm2   & 0.250 & 0.310 & \textbf{0.249} & 0.310 & 0.252 & \textbf{0.308} & 0.250 & 0.310 & 0.250 & 0.310 \\
Weather & 0.243 & 0.273 & 0.244 & 0.273 & \textbf{0.238} & \textbf{0.273} & 0.243 & 0.278 & 	0.243 & 0.278 \\
\bottomrule
\end{tabular}
\end{table}

\subsection{Wavelet Depth and Wavelet Pool}
\label{app:wavelet-depth-pool}

We next ablate two design choices in the wavelet view generator: the decomposition depth $J$ and the wavelet sampling strategy. 
The depth $J$ determines the scale at which the signal is split into the shared low-frequency approximation and the editable detail subspace. 
A shallow decomposition leaves more local variation in the approximation, while a deeper decomposition makes the teacher view more aggressively coarse. 
We compare $J\in\{2,3,4\}$ using the same DINO objective and backbone.

Table~\ref{tab:wavelet-depth-ablation-appendix} shows that intermediate depth provides the most stable performance. 
With $J=2$, the easy and hard views remain too close in scale, weakening the self-distillation signal. 
With $J=4$, the approximation becomes overly coarse on several datasets, causing the teacher target to discard useful local structure. 
The default $J=3$ provides the best overall trade-off and is used throughout the main experiments.

We also evaluate the effect of wavelet-basis sampling. 
Specifically, we compare a fixed-basis setting, a shared-sampled setting in which the easy and hard views use the same randomly sampled wavelet, and the default independent-sampling setting in which $w^{\mathrm{easy}}$ and $w^{\mathrm{hard}}$ are drawn independently from the wavelet pool. 
Table~\ref{tab:wavelet-pool-ablation-appendix} shows that independent sampling is the most robust overall. 
Using a fixed basis performs competitively on some datasets but is less stable, suggesting that the encoder can overfit to basis-specific phase or boundary behavior. 
Shared sampling improves over a fixed basis by adding wavelet diversity, while independent sampling further improves robustness by introducing mild cross-basis leakage between the easy and hard views. 
This supports the default design of sampling the two view bases independently.

\begin{table}[t]
\centering
\caption{
Wavelet-depth ablation using linear probe. 
Each cell reports MSE/MAE averaged over horizons $\{96,192,336,720\}$. 
Lower is better; best results are in \textbf{bold}.
}
\label{tab:wavelet-depth-ablation-appendix}
\scriptsize
\setlength{\tabcolsep}{3.0pt}
\begin{tabular}{lcccccc}
\toprule
 & \multicolumn{2}{c}{$J=2$} & \multicolumn{2}{c}{$J=3$} & \multicolumn{2}{c}{$J=4$} \\
\cmidrule(lr){2-3}\cmidrule(lr){4-5}\cmidrule(lr){6-7}
\textbf{Dataset} & MSE & MAE & MSE & MAE & MSE & MAE \\
\midrule
ETTh1   & 0.426 & 0.429 & \textbf{0.416} & \textbf{0.423} & 0.438 & 0.440 \\
ETTh2   & 0.370 & 0.395 & \textbf{0.347} & \textbf{0.385} & 0.361 & 0.391 \\
ETTm1   & 0.361 & 0.386 & \textbf{0.364} & \textbf{0.386} & 0.362 & 0.385 \\
ETTm2   & 0.255 & 0.312 & \textbf{0.252} & \textbf{0.308} & 	0.248 & 0.308 \\
Weather & 0.238 & 0.272 & \textbf{0.238} & \textbf{0.273} & 	0.239 & 0.273 \\
\bottomrule
\end{tabular}
\end{table}

\begin{table}[t]
\centering
\caption{
Wavelet-pool and basis-sampling ablation (Linear probe). 
Fixed uses a single wavelet basis for both views. 
Shared sampled draws one wavelet from $\mathcal{P}$ and uses it for both easy and hard views. 
Independent sampled is the WINO-TS default, drawing $w^{\mathrm{easy}}$ and $w^{\mathrm{hard}}$ independently. 
Each cell reports MSE/MAE averaged over horizons $\{96,192,336,720\}$. 
Lower is better; best results are in \textbf{bold}.
}
\label{tab:wavelet-pool-ablation-appendix}
\scriptsize
\setlength{\tabcolsep}{5pt}
\begin{tabular}{lcccccc}
\toprule
& \multicolumn{2}{c}{\shortstack{\textbf{Fixed}\\\textbf{basis}}} 
& \multicolumn{2}{c}{\shortstack{\textbf{Shared}\\\textbf{sampled}}} 
& \multicolumn{2}{c}{\shortstack{\textbf{Independent}\\\textbf{sampled}}} \\
\cmidrule(lr){2-3} \cmidrule(lr){4-5} \cmidrule(lr){6-7}
\textbf{Dataset} & MSE & MAE & MSE & MAE & MSE & MAE \\
\midrule
ETTh1   & 0.429 & 0.431 & 0.420 & 0.429 & \textbf{0.416} & \textbf{0.423} \\
ETTh2   & 0.369 & 0.394 & 0.363 & 0.391 & \textbf{0.347} & \textbf{0.385} \\
ETTm1   & \textbf{0.362} & 0.387 & 0.363 & 0.387 & 0.364 & \textbf{0.386} \\
ETTm2   & \textbf{0.247} & \textbf{0.308} & 0.275 & 0.328 & 0.252 & \textbf{0.308} \\
Weather & 0.243 & 0.279 & 0.243 & 0.279 & \textbf{0.238} & \textbf{0.273} \\
\bottomrule
\end{tabular}
\end{table}

\subsection{View-Pairing Ablation}
\label{app:view-pairing-ablation}

The default WINO-TS configuration follows an asymmetric DINO-style
view-pairing rule. The teacher processes only the easy views, whereas
the student processes both easy and hard views. Every student view is
matched to every easy teacher view, except that the teacher and
student outputs corresponding to the identical easy-view realization
are not paired. Thus, both student easy and student hard views
contribute to the default objective.

We compare this default rule against two alternative pairing
strategies. The \emph{same-view} variant additionally includes the
matching easy-teacher/easy-student pairs that are excluded by the
default objective. The \emph{symmetric} variant allows both the
teacher and student to process easy and hard views and applies
cross-view matching across non-identical view pairs.

Table~\ref{tab:view-pairing-ablation-appendix} shows that the default
asymmetric DINO-style pairing provides the most stable performance
across the evaluated datasets. Including identical easy--easy pairs
does not yield consistent gains, suggesting that direct same-view
matching can weaken the desired cross-view invariance pressure. The
fully symmetric formulation is competitive on ETTm1 and ETTm2 but is
less stable on ETTh1, ETTh2, and Weather.

These results indicate that the benefit of WINO-TS depends not only
on generating wavelet-domain views but also on assigning distinct
roles to the two networks: the teacher produces targets only from
easy views, while the student learns from both easy and hard views
through non-trivial cross-view matching.

\begin{table}[t]
\centering
\caption{
View-pairing ablation using linear probing.
\emph{Default} denotes the WINO-TS pairing rule: the teacher processes
easy views, the student processes both easy and hard views, and every
student view is matched to every teacher easy view except the
identical easy-view index.
\emph{Same-view} additionally includes the matching easy--easy pairs.
\emph{Symmetric} allows both networks to process easy and hard views.
Each cell reports MSE/MAE averaged over horizons
$\{96,192,336,720\}$. Lower is better; best results are in
\textbf{bold}.
}
\label{tab:view-pairing-ablation-appendix}
\scriptsize
\setlength{\tabcolsep}{4pt}
\begin{tabular}{lcccccc}
\toprule
& \multicolumn{2}{c}{\textbf{Default}} 
& \multicolumn{2}{c}{\textbf{Same-view}} 
& \multicolumn{2}{c}{\textbf{Symmetric}} \\
\cmidrule(lr){2-3} \cmidrule(lr){4-5} \cmidrule(lr){6-7}
\textbf{Dataset} & MSE & MAE & MSE & MAE & MSE & MAE \\
\midrule
ETTh1   & \textbf{0.416} & \textbf{0.423} & 0.435 & 0.434 & 0.420 & 0.430 \\
ETTh2   & \textbf{0.347} & \textbf{0.385} & 0.363 & 0.391 & 0.364 & 0.392 \\
ETTm1   & 0.364 & 0.386 & \textbf{0.361} & \textbf{0.385} & 0.362 & \textbf{0.385} \\
ETTm2   & 0.252 & 0.308 & \textbf{0.248} & 0.309 & \textbf{0.248} & \textbf{0.307} \\
Weather & \textbf{0.238} & \textbf{0.273} & 0.242 & 0.279 & 0.239 & 0.275 \\
\bottomrule
\end{tabular}
\end{table}

\subsection{Seed Robustness}
\label{app:seed-robustness}

We evaluate WINO-TS-LP across five random seeds, $\{42,777,1773,2024,3407\}$. The results are depicted in Table \ref{app:seed-robustness}. The standard deviations are small across the tested seeds, indicating limited run-to-run variability for the reported linear-probe setting.

\begin{table}[t]
\centering
\caption{
\textbf{Seed robustness of WINO-TS LP} (DINO pre-training, TimeMixer backbone, linear probe).
In-domain forecasting MSE/MAE averaged over horizons $\{96,192,336,720\}$,
reported as mean $\pm$ std over 5 seeds ($\{42, 777, 1773, 2024, 3407\}$). Lower is better.
}
\label{tab:seed-robustness}
\scriptsize
\setlength{\tabcolsep}{6pt}
\begin{tabular}{lcc}
\toprule
\textbf{Dataset} & \textbf{MSE} & \textbf{MAE} \\
\midrule
ETTh1   & $0.417 \pm 0.001$ & $0.424 \pm 0.001$ \\
ETTh2   & $0.350 \pm 0.004$ & $0.387 \pm 0.002$ \\
ETTm1   & $0.361 \pm 0.003$ & $0.385 \pm 0.001$ \\
ETTm2   & $0.248 \pm 0.002$ & $0.308 \pm 0.002$ \\
Weather & $0.238 \pm 0.001$ & $0.272 \pm 0.001$ \\
\bottomrule
\end{tabular}
\end{table}

\begin{table}[h]
\centering
\caption{
\textbf{Multi-view ablation} with a TimeMixer backbone and linear
probing. The default uses $Q{=}1$ easy view and $V{=}1$ hard view;
the alternatives use $Q{=}2$ and $V\in\{2,4,6\}$. Each view uses an
independently sampled wavelet basis and, for hard views, independently
sampled noise. Results are MSE/MAE averaged over horizons
$\{96,192,336,720\}$; lower is better and best results are in
\textbf{bold}.
}
\label{tab:app-hardcrop-sweep}
\scriptsize
\setlength{\tabcolsep}{3pt}
\begin{tabular}{lcccccccc}
\toprule
\textbf{Dataset} & \multicolumn{2}{c}{\begin{tabular}[t]{@{}c@{}}\textbf{WINO-TS}\\ (\textbf{$Q{=}1,V{=}1$})\end{tabular}} & \multicolumn{2}{c}{\textbf{$Q{=}2,V{=}2$}} & \multicolumn{2}{c}{\textbf{$Q{=}2,V{=}4$}} & \multicolumn{2}{c}{\textbf{$Q{=}2,V{=}6$}} \\
\cmidrule(lr){2-3}\cmidrule(lr){4-5}\cmidrule(lr){6-7}\cmidrule(lr){8-9}
        & \B{MSE} & \B{MAE} & \B{MSE} & \B{MAE} & \B{MSE} & \B{MAE} & \B{MSE} & \B{MAE} \\
\midrule
ETTh1   & \B{\textbf{0.416}} & \B{\textbf{0.423}} & \B{0.433} & \B{0.433} & \B{0.429} & \B{0.431} & \B{0.429} & \B{0.430} \\
ETTh2   & \B{\textbf{0.347}} & \B{\textbf{0.385}} & \B{0.363} & \B{0.391} & \B{0.362} & \B{0.391} & \B{0.362} & \B{0.390} \\
ETTm1   & \B{0.364} & \B{0.386} & \B{\textbf{0.361}} & \B{\textbf{0.385}} & 0.362 & 0.385 & 0.362 & 0.386 \\
ETTm2   & \B{0.252} & \B{\textbf{0.308}} & \B{\textbf{0.247}} & \B{0.308} & \B{0.248} & \B{0.309} & 0.247 & 0.308 \\
Weather & \B{\textbf{0.238}} & \B{\textbf{0.273}} & \B{0.240} & \B{0.276} & \B{0.239} & \B{0.275} & \B{0.239} & \B{0.275} \\
\bottomrule
\end{tabular}
\end{table}

\FloatBarrier


\end{document}